\documentclass[runningheads]{llncs}

\usepackage{eccv}

\usepackage{algorithm}
\usepackage{algorithmic}
\usepackage[utf8]{inputenc} 
\usepackage{booktabs}       
\usepackage{amsfonts}       
\usepackage{nicefrac}       
\usepackage{microtype}      
\usepackage{xcolor}         
\usepackage{dsfont}
\usepackage{multirow}
\usepackage{newfloat}
\usepackage{listings}
\usepackage{slashbox}
\usepackage{diagbox}
\usepackage{tikz}
\usepackage{eccvabbrv}
\usepackage{xcolor,colortbl}
\usepackage{graphicx}
\usepackage{booktabs}
\usepackage{bbding}
\usepackage[accsupp]{axessibility}  

\usepackage{hyperref}

\usepackage{orcidlink}
\usepackage{wrapfig,booktabs}

\begin{document}

\title{Category Adaptation Meets Projected Distillation in Generalized Continual Category Discovery}

\titlerunning{Category Adaptation Meets Projected Distillation}

\author{
Grzegorz~Rypeść\thanks{Corresponding author, email: grzegorz.rypesc.dokt@pw.edu.pl}\inst{1,2}\orcidlink{0000-0001-8170-3282} \and
Daniel~Marczak\inst{1,2}\orcidlink{0000-0002-6352-9134} \and
Sebastian~Cygert\inst{1,3}\orcidlink{0000-0002-4763-8381} \and \\
Tomasz~Trzciński\inst{1,2,4}\orcidlink{0000-0002-1486-8906} \and
Bartłomiej~Twardowski\inst{1,5,6}\orcidlink{0000-0003-2117-8679}
}

\authorrunning{G.~Rypeść et al.}

\institute{
IDEAS~NCBR \and
Warsaw~University~of~Technology \and 
Gdańsk~University~of~Technology \and 
Tooploox \and
Autonomous~University of~Barcelona \and 
Computer~Vision~Center
}

\maketitle

\begin{abstract}

Generalized Continual Category Discovery (GCCD) tackles learning from sequentially arriving, partially labeled datasets while uncovering new categories. Traditional methods depend on feature distillation to prevent forgetting the old knowledge. However, this strategy restricts the model's ability to adapt and effectively distinguish new categories. To address this, we introduce a novel technique integrating a learnable projector with feature distillation, thus enhancing model adaptability without sacrificing past knowledge. The resulting distribution shift of the previously learned categories is mitigated with the auxiliary category adaptation network. We demonstrate that while each component offers modest benefits individually, their combination -- dubbed CAMP (Category Adaptation Meets Projected distillation) -- significantly improves the balance between learning new information and retaining old. CAMP exhibits superior performance across several GCCD and Class Incremental Learning scenarios. The code is available on \href{https://github.com/grypesc/CAMP}{Github}.

%

\end{abstract}

\section{Introduction}

Traditional machine learning models provide remarkable performances across various applications. However, they typically operate within the closed-world scenario, which assumes that all the tasks are well-defined and knowledge necessary to solve them is given upfront.
Unfortunately, those assumptions are far from real-life applications, where models encounter dynamic streams of ever-changing data that evolves and may contain semantics the models have never encountered before. This variability severely impacts their performance and robustness. To study these problems from a unified perspective, many recent works combine Generalized Category Discovery~\cite{vaze2022generalized} (GCD) setting with Continual Learning\cite{li2017learning, masana2022class} (CL). 
Resulting combinations~\cite{zhao2023incremental, kim2023proxy, wu2023metagcd, zhang2022grow} (to which we refer as Generalized Continual Category Discovery - GCCD), consider a sequence of tasks, in which the method has to continuously handle new data, discover novel categories and maintain good performance on previous tasks by overcoming catastrophic forgetting~\cite{french1999catastrophic}. 
A popular trend in the field of discovering categories is to regularize the model with feature distillation~\cite{kim2023proxy, roy2022class, Joseph2022NovelClassDisco} (FD) to alleviate forgetting. However, as pointed out by~\cite{gomezvilla2022continually}, it decreases the plasticity of the model by restricting changes (drift) of the categories' distributions in the feature space. In this work, we show that in order to enable learning robust representations of new data, it is necessary to allow distributions of past categories to drift. Drawing inspiration from recent studies in representation learning for enhanced transferability~\cite{bordes2023guillotine, sariyildizKAL23Trex}, we incorporate projected representation in a distillation-based regularization for continual learning. Compared to standard FD, this method results in improved plasticity. However, it increases forgetting as it lets old categories drift even more.

\begin{wrapfigure}[26]{r}{0.52\textwidth}
    \centering
    \includegraphics[width=0.52\textwidth]{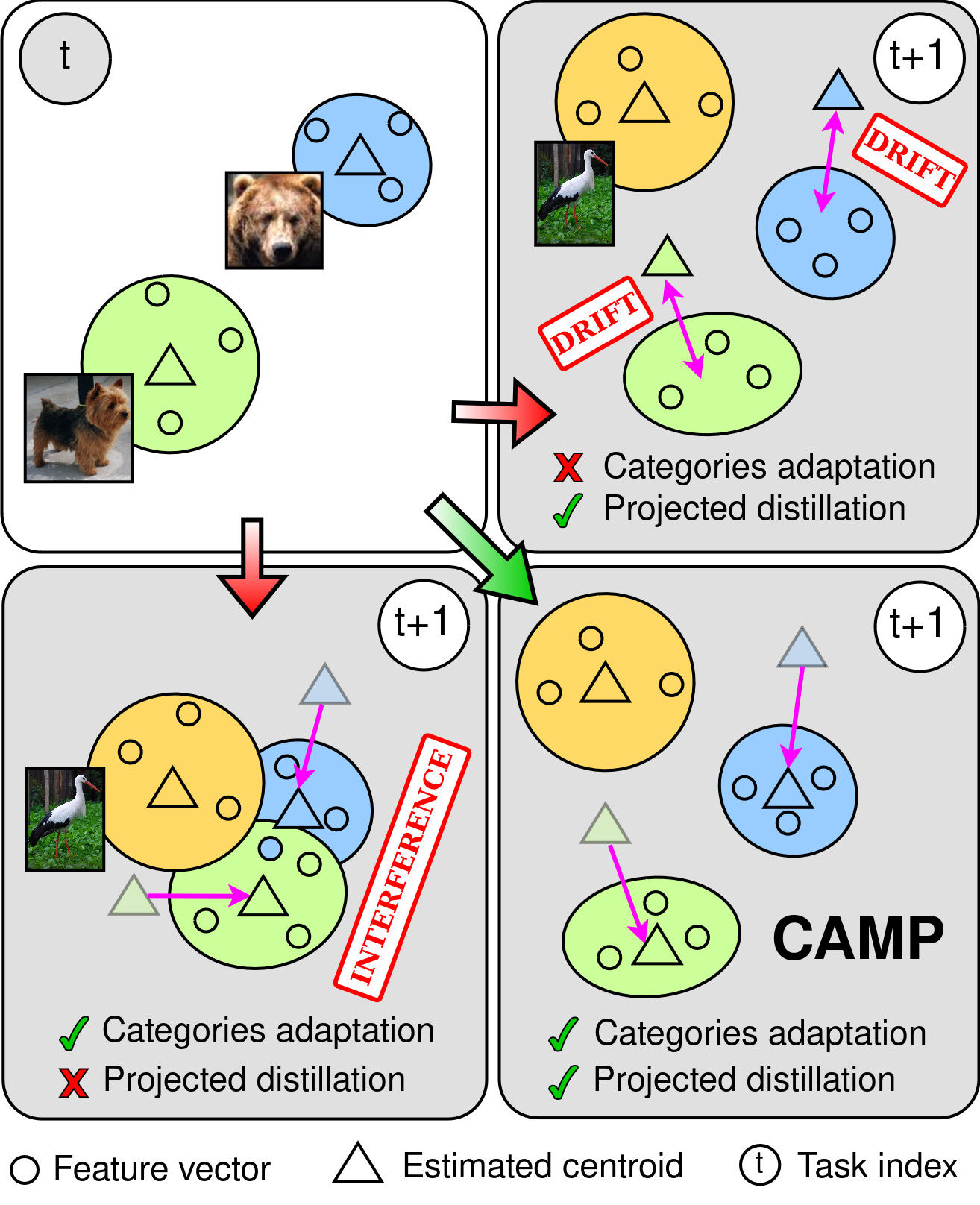}

    \caption{While knowledge distillation through a projector and category adaption are decent on their own, CAMP combines them to achieve great results without the need of exemplars.}
    \label{fig:teaser}
\end{wrapfigure}

To address this issue, we represent categories as centroids in the latent space of the feature extractor and train an auxiliary network to predict the drift of centroids, which we call category adaptation. This enables the model to recover the actual positions of old clusters and improve the overall accuracy of the Nearest Centroid Classifier (NCC). Similarly, works~\cite{iscen2020memory, yu2020semantic} analyze the semantic drift of old categories in closed-world, Continual Learning scenarios. Here, our key observation is that, while centroid adaptation demonstrates overall solid performance, its true effectiveness is realized only through \underline{combining} it with feature distillation using a projection network, as depicted in Fig.~\ref{fig:teaser}. That allows the technique to excel. Building upon these findings, we propose to leverage this combined effect for GCCD by introducing a method called CAMP -- \underline{C}ategory \underline{A}daptation \underline{M}eets \underline{P}rojected distillation. CAMP presents state-of-the-art performance across GCCD scenarios and exemplar-free Class Incremental Learning (CIL). 

To summarize, the main contributions of our work are as follows. \textbf{1:} We study the interplay between knowledge distillation and category adaptation for the problem of Generalized Continual Category Discovery (GCCD). We demonstrate that they address distinct or even opposing challenges when considered separately. However, when combined, they complement each other significantly, reduce forgetting, and improve overall results.
\textbf{2:} Based on those insights, we propose a novel method for generalized continual category discovery dubbed CAMP. Built on GCD~\cite{vaze2022generalized}, CAMP improves it for continual learning by adapting categories between tasks and performing knowledge distillation with two separate projectors.
\textbf{3:} We evaluate our method on numerous GCCD and Class Incremental Learning (CIL) datasets. With a series of experiments, we show that CAMP achieves state-of-the-art results with and without exemplars in different scenarios.

\section{Related work}

\textbf{Continual learning (CL)} is a setting where an agent learns a sequence of tasks with access to the data from the current task only. The goal is to achieve high performance on all the tasks from the sequence. Regularization-based approaches, such as EWC~\cite{kirkpatrick2017overcoming} or MAS~\cite{AljundiBERT18} penalize changes to important neural network parameters. 
Alternatively, it is possible to regularize neuron activations as in LwF~\cite{li2017learning}, DER~\cite{yan2021dynamically} or PODNet~\cite{DouillardCORV20} by using distillation techniques.
However, even with feature distillation, the features from old classes will change, which causes catastrophic forgetting. \cite{yu2020semantic} tries to predict these changes by approximating the drift of old classes based on the drift of current task data. ~\cite{iscen2020memory} learns a feature adaptation network that translates old features into those learned on new tasks. In our work, we show that such an approach may not necessarily work well when going beyond the closed-world assumption under which CL usually operates.

\noindent\textbf{Novel Class Discovery} (NCL)~\cite{han2019learning,han2020automatically,zhong2020openmix,HanREVZ22,FiniSLZN021,ZhongFRL0S21} aims to discover unknown classes from unlabeled data, thus it relaxes the closed-world assumption. NCL typically assumes that the class space of unlabeled data is completely disjoint with the labeled (training) set. This assumption was removed by Generalized Category Discovery (GCD)~\cite{vaze2022generalized, wen2023parametric},
where a training dataset consists of two parts: known classes where only a fraction of samples are labeled and novel classes which are unlabeled. The goal is to categorize all images from known and novel classes, without any additional information regarding their novelty. 
Typical solutions in this area include unsupervised clustering~\cite{han2019learning,zhao2023learning}, contrastive learning~\cite{vaze2022generalized,zhao2023learning,wen2023parametric}, pseudo-labeling~\cite{wen2023parametric} or pairwise similarities~\cite{zhang2023promptcal}.

\noindent\textbf{Continual classes discovery} combines the NCL and CL fields. 
In~\cite{roy2022ClassIncr}, learning of the novel classes is framed incrementally: that is, in the first stage, the model learns from supervised data, and in the following stages, novel classes are learned without any labels. A similar setting was explored in~\cite{Joseph2022NovelClassDisco} with the solution that uses pseudo-latent supervision, feature
distillation, and a mutual information-based regularizer. 
However, their setting assumes no class overlap between the initial and incremental training stages, which was later relaxed by various to which we refer as Generalized Continual Category Discovery - GCCD~\cite{zhao2023incremental, kim2023proxy, wu2023metagcd, zhang2022grow}.
Grow and Merge (GM)\cite{zhang2022grow}  keeps two models: a static one that maintains knowledge of old classes and a dynamic one trained using self-supervision to discover novel classes. Both models are then merged into one when the new task arrives.
In IGCD~\cite{zhao2023incremental} a non-parametric classifier and a small buffer of exemplars from previous tasks to decrease forgetting. PA~\cite{kim2023proxy} combines exemplars and proxy anchors~\cite{kim2020proxy} known from deep metric learning.
Contrary to previous approaches~\cite{zhang2022grow,zhao2023incremental,kim2023proxy} we focus on the most challenging exemplar-free setting. MetaGCD~\cite{wu2023metagcd} needs to store the data from the first task, which we avoid, by proposing a purely continual model.

\section{Method}
\label{sec:method}
\subsection{Problem Setting}

In Generalized Continual Category Discovery models are trained sequentially on partially labeled tasks containing known and novel categories.
Formally, we consider a scenario where a dataset $D$ is partitioned into disjoint subsets, each represented by a task: $D=T^1 \cup T^2 \cup \dots \cup T^{N}$, with $N$ denoting the total number of tasks. Each task $T^t$ comprises two subsets: $T^t = T_L^t \cup T_U^t$, wherein 
$T_L^t = \{(x, y) \mid x \in \mathcal{X}^t_L, y \in \mathcal{Y}^t\}$
denotes the labeled set of known classes and
$T_U^t = \{x \mid x \in \mathcal{X}_U^t\}$
encompasses unlabeled data samples from both known and novel classes.

Within the GCCD framework, each method engages in sequential learning from a stream of tasks. We assume that, at task $t$, each method comprises a feature extractor $\mathcal{F}^t:\mathcal{X}^t \rightarrow \mathcal{Z}^t$ and a classifier $C^t:\mathcal{Z}^t \rightarrow \mathcal{Y}^t$. Here, $\mathcal{Z}^t$ signifies the latent space generated by the feature extractor trained on the task $t$. The GCCD model aims to learn to recognize categories of a given task $T^t$ while also maintaining the ability to recognize the categories from previous tasks $T^1 \cup T^2 \cup \dots \cup T^{t-1}$. The main challenge is to prevent catastrophic forgetting on previous tasks while maintaining high plasticity that allows a method to learn new categories on new data.

\subsection{Motivation}
\label{drift-visualization}
\begin{figure}[ht!]
    \centering
    \includegraphics[width=1.0\textwidth]{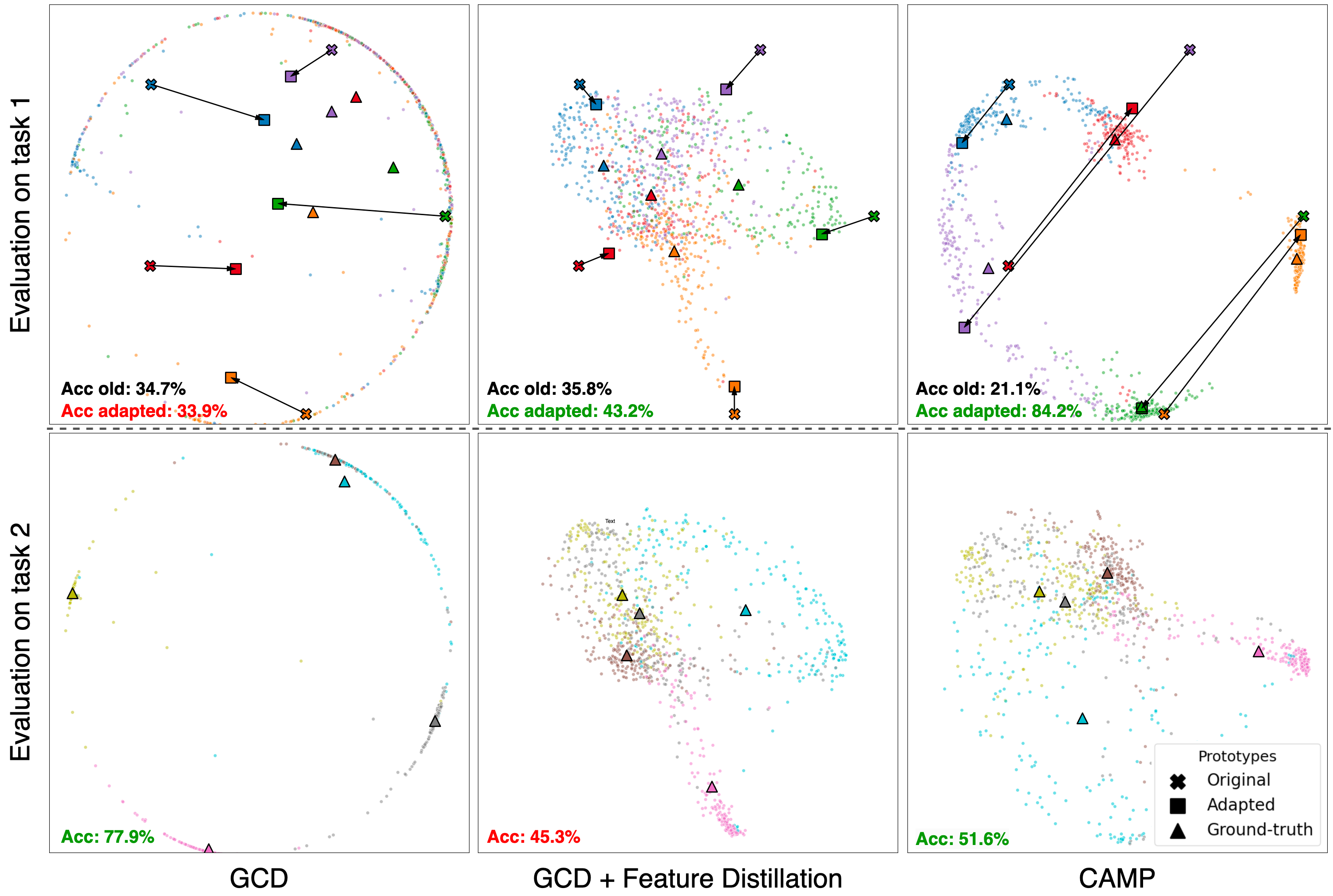}
    \caption{CAMP utilizes a projected knowledge distillation, resulting in a predictable latent space drift. The drift is revertible via centroid adaptation (black arrows), maintaining high performance on the first task. GCD and GCD with feature distillation fail to prevent forgetting, resulting in a drift that is difficult to predict. This decreases their performance. We report the nearest centroid classification accuracy using stored (Acc old) and adapted centroids (Acc adapted) after training on the second task.} 
    \label{fig:drift_viz}
\end{figure}

In this section, we demonstrate the motivation for our method on a simple experiment performed on ten categories of the CUB200 dataset split into two tasks. We train a neural network with its latent space bottlenecked to two dimensions (see Supplementary Material B for experimental details). In Fig.~\ref{fig:drift_viz}, we present a visualization of the latent space after training on the second task.
We observe that naive fine-tuning of GCD (left) causes catastrophic forgetting and results in poor performance on the first task.
Training with feature distillation (middle)~\cite{roy2022class, kim2023proxy, Joseph2022NovelClassDisco} to some extent alleviates the forgetting and makes the drift of the centroids partially reversible via centroid adaptation (black arrows). However, feature distillation introduces rigid regularization that limits the model's ability to acquire new knowledge and results in low performance on the second task. These results highlight the need for a mechanism that relaxes anti-forgetting regularization constraints while alleviating forgetting. Our proposed CAMP method (right) is able to address these issues by introducing a learnable projector, which makes the regularization less rigid and allows the distribution to drift more, resulting in better accuracy on the second task. At the same time, it causes the drift to be easily predictable, thus enabling our centroid adaptation to estimate ground truth positions of centroids, leading to great accuracy on the first task. Existing feature adaptation methods~\cite{yu2020semantic, iscen2020memory} do not utilize projected distillation. Therefore, their drift is more challenging to predict (middle). Additionally, our method is more memory efficient than ~\cite{iscen2020memory} as we store only a single centroid instead of many features per class and we don't require exemplars.

\subsection{CAMP}
\label{Method:representation}
\begin{figure*}[h]
    \centering
    \includegraphics[width=1.0\textwidth]{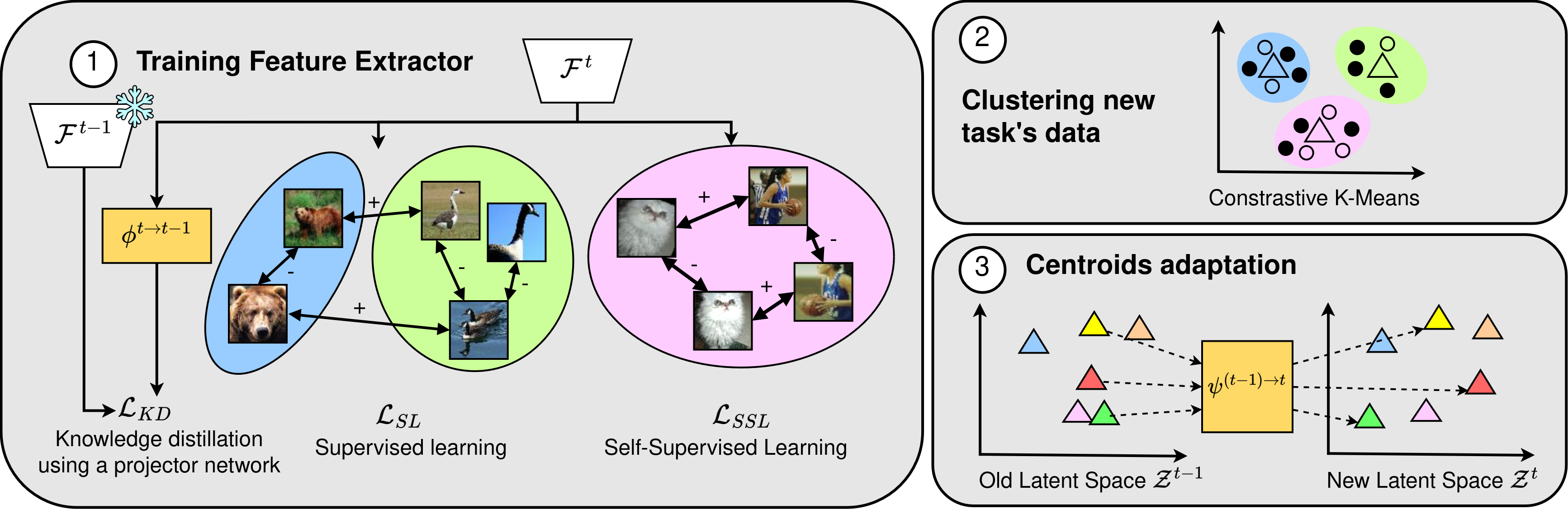}
    \caption{The training procedure of CAMP consists of three stages: (1) We train the feature extractor in a semi-supervised manner using distillation through a learnable projector. (2) We obtain centroids of known and novel classes in the current task using constrastive K-Means algorithm. (3) We update memorized centroids of old categories to alleviate forgetting.}
    \label{fig:method}
\end{figure*}

The training procedure for CAMP within each task is divided into three consecutive phases: (1) feature extractor training, (2) clustering on a new task, and (3) adaptation of centroids representing categories of past classes. An overview is depicted in Fig.~\ref{fig:method}. The first phase focuses on training $\mathcal{F}^t$ to transform images into meaningful features that effectively distinguish between different classes. The second phase is dedicated to identifying novel classes. The third phase, centroid adaptation, is responsible for updating the centroids of previous tasks by estimating the drift caused by updating the feature extractor. Detailed explanations of each phase will follow in the subsequent three subsections.

\subsection{Phase 1: Feature extractor training}

Our method for training feature extractors consists of three components: self-supervised learning, supervised learning, and knowledge distillation. The first two components allow the use of labeled and unlabeled data for the training feature extractor, while knowledge distillation is used for regularization and preventing forgetting. All are depicted in Fig.~\ref{fig:method}, left rounded rectangle.

\noindent\textbf{\textit{Representation learning}}
To learn the representations of the data from the new task, we follow SimGCD~\cite{wen2023parametric} and combine self-supervised and supervised learning. We use self-supervision on all data from known and novel categories $T^U_t$. We combine SimCLR~\cite{chen2020simple} loss and cross-entropy loss between predictions and pseudo-labels: $\mathcal{L}_{SSL} = \mathcal{L}_{SimCLR} + \mathcal{L}_{pseudo}$. For labeled data $T^L_t$ we use supervised learning loss that combines SupCon~\cite{khosla2021supervised} loss and cross-entropy loss between predicted and ground-truth labels: $\mathcal{L}_{SL} = \mathcal{L}_{SupCon} + \mathcal{L}_{CE}$. A detailed formal description of each loss component is provided in Supplementary Material~A.

\noindent\textbf{\textit{Knowledge distillation}}
We use distillation to alleviate forgetting in continual learning scenarios. Following state-of-the-art exemplar-free continual representation learning methods~\cite{gomezvilla2022continually, fini2022cassle}, we use a learnable MLP projector $\phi^{t \rightarrow t-1}: \mathcal{Z}^t \rightarrow \mathcal{Z}^{t-1}$ that maps the features learned on the current task back to the old latent space. In our case, this approach improves the trade-off between plasticity and stability. Additionally, it ensures there is a mapping between old ($t-1$) and new ($t$) latent space, which can be later inverted during category adaptation.

Following distillation-based continual learning methods~\cite{li2017learning, gomezvilla2022continually, fini2022cassle}, we freeze previously trained feature extractor $\mathcal{F}^{t-1}$ and regularize currently trained feature extractor $\mathcal{F}^t$ with the outputs of $\mathcal{F}^{t-1}$:
\begin{equation}
\label{eq:knowledge_distillation}
\mathcal{L}_{KD} = \sum_{i \in B} || \phi^{t \rightarrow t-1}(\mathcal{F}^{t}(x_i)) - \mathcal{F}^{t-1}(x_i) ||^2.
\end{equation}
where $B$ is a batch of data.
This form of distillation does not require labels. Therefore, all the data from $T_U^t$ can be used for regularization.

The final loss function for feature extractor training is defined as follows:
\begin{equation}
\label{eq:total_loss}
    \mathcal{L} = (1-\alpha)((1-\beta)\mathcal{L}_{SSL} + \beta \mathcal{L}_{SL}) + \alpha \mathcal{L}_{KD},
\end{equation}
where $\alpha, \beta \in [0,1]$ are hyperparameters defining contribution of regularization and supervision respectively.

\subsection{Phase 2: Clustering task's data}

The purpose of clustering the new task's data $T^t$ is to find a centroid of each known and novel class in the learned representation space of the feature extractor $\mathcal{F}^t$. Such centroid is the class representation in latent space, later used for classification with the Nearest Centroid Classifier (NCC) during inference time. In order to find class centroids for CAMP, we use a semi-supervised k-means algorithm introduced in \cite{vaze2022generalized}. We use labeled data samples of known classes ($T_L^t$) to initialize known classes' centroids. We obtain the set of remaining centroids (equal to the number of novel classes) from the unlabeled data $T_U^t$ using the k-means++~\cite{arthur2007k} algorithm, with the constraint that they must be close to the centroids of the labeled ones from $T_L^t$. To determine the number of classes in a task, we utilize a popular elbow method as in\cite{han2019learning, vaze2022generalized}. Thus, for different numbers of K-Means clusters, we measure clustering accuracy on the labeled data and set the number of clusters to the one that yields the highest accuracy.

\subsection{Phase 3: Centroid adaptation}

Although regularization methods decrease forgetting, they still allow for a drift of ground truth representations of old classes (as discussed in Sec.~\ref{drift-visualization}). To recover from this issue, we propose to train auxiliary centroid adaptation network $\psi^{(t-1) \rightarrow t}$ to predict the drift of data representations from the previous task's latent space ($t-1$) to the current one ($t$). To train it, we use all the data available in the current task. Exemplars can also be used for this purpose but are not necessary. We examine the architecture of $\psi^{(t-1) \rightarrow t}$ in Sec.~\ref{sec:method_analysis}.
We train $\psi^{(t-1) \rightarrow t}$ by minimizing the MSE loss:
\begin{equation}
\label{eq:prototypes_adaptation_network}
\mathcal{L}_{PA} = \sum_{i \in B} || \mathcal{F}^{t}(x_i) - \psi^{(t-1) \rightarrow t} (\mathcal{F}^{t-1}(x_i)) || ^ 2.
\end{equation}
When $\psi^{(t-1) \rightarrow t}$ is trained, we use it to calculate the updated centroids $p^{t}_i = \psi^{(t-1) \rightarrow t}(p^{t-1}_i)$ where $p^t_i$ is a centroid of class $i$ calculated after task $t$. This is presented in Fig.~\ref{fig:teaser}, lower right.

\section{Experiments}

\subsection{Experimental setup}
We evaluate all methods on five datasets: CIFAR100~\cite{krizhevsky2009learning}, Stanford Cars~\cite{stanford_cars}, CUB200~\cite{wah2011caltech}, FGVCAircraft~\cite{maji2013fine} and DomainNet~\cite{peng2019moment}. We split datasets into five equal tasks, with the exception of Stanford Cars (4 tasks) and DomainNet (6 tasks). Following~\cite{rypesc2024divide} we set a different domain for each task of DomainNet, to simulate domain shifts. Following~\cite{vaze2022generalized, wen2023parametric}, for each dataset, we set the ratio of known to novel classes as 4:1. For known classes, we set the ratio of labeled data samples to unlabeled data as 1:1. As the feature extractor $\mathcal{F}$ for all methods we use ViT-small~\cite{dosovitskiy2020vit} model pretrained with DINO~\cite{caron2021emerging} on ImageNet~\cite{deng2009imagenet}. Following GCD~\cite{vaze2022generalized} we freeze the first 11 blocks for all methods. We train all methods for 100 epochs in each task using AdamW optimizer with a starting learning rate of $0.1^3$ and cosine annealing scheduling. The exception is PA method, for which we set it to $0.1^4$ but increase the learning rate for proxies 100 times following the original work. As the distiller $\phi$ we utilize a 2-layer MLP network consisting of 384 neurons and ReLU activation. As the adapter $\psi$, we utilize a linear layer consisting of 384 neurons. For CAMP without exemplars, we set $\alpha$ to 0.5; for CAMP with exemplars, we use $\alpha=0.1$. 

\noindent\textbf{Selection of baselines}
For the simplest baseline, we fine-tune GCD~\cite{vaze2022generalized} method and couple it with popular CL regularization-based methods: (1) weight regularization method - EWC~\cite{kirkpatrick2017overcoming} (GCD+EWC); (2) feature distillation method~\cite{li2017learning, gomezvilla2022continually} (GCD+FD). In GCD+FD we distill features $\mathcal{F}(x)$. Moreover, we combine GCD with experience replay~\cite{riemer2018learning} (GCD+ER), where for incremental tasks, we add a buffer that stores random images of past known classes. For all the above methods, we perform semi-supervised k-means~\cite{vaze2022generalized}  algorithm to find centroids and utilize the NCC classifier for classification. Additionally, we use SimGCD~\cite{wen2023parametric}, which improves GCD by utilizing a learnable classifier and adding additional cross-entropy-based losses for training it. We also test MetaGCD\cite{wu2023metagcd} that utilizes meta-learning and PA~\cite{kim2023proxy} method, which trains the feature extractor using supervised proxy anchor loss, maintains a buffer of exemplars, and uses feature distillation to alleviate forgetting. Lastly, we evaluate IGCD~\cite{zhao2023incremental} that utilizes a density-based support set selection and replay buffer to mitigate forgetting. We describe all methods in detail in the Supplementary Material C.


\noindent\textbf{Evaluation protocol}
We assess the performance of each method upon completion of the final task using the test set for each task. Our primary evaluation metrics encompass final task-agnostic (TAg) accuracy~\cite{chaudhry2018riemannian} (where the model does not know task-id during the inference), average forgetting~\cite{chaudhry2018riemannian} after last task and plasticity defined as $\frac{1}{N-1}\Sigma_{n=2}^{N} A_n$, where $A_n$ denotes accuracy measured on n-th task after training on n-th task. Following~\cite{vaze2022generalized, wen2023parametric}, we calculate them separately for known, novel, and all classes to get more insight into the performance of methods. We will release the code upon the acceptance of the manuscript.

\subsection{Comparison to baselines}

We compare CAMP to other methods in terms of accuracy in Tab.~\ref{tab:main_results}. CAMP outperforms existing methods by a large margin in most scenarios, achieving the best results on CIFAR100, Stanford Cars, and CUB200 with and without exemplars, \eg on CUB200, CAMP with no exemplars achieves 8.6\%, 9.3\%,  6.1\% points better all, known and novel accuracy than the second-best method GCD+EWC. However, results are not that consistent on FGVCAircraft and DomainNet. On DomainNet GCD+FD method achieves 10.0\% points better novel accuracy than CAMP, yet still, our method has 4.0\% points better all accuracy. We can also study the impact of exemplars. Methods that utilize exemplars outperform exemplar-free competitors at the cost of additional memory complexity. Exemplars drastically improve GCD method, on CUB200 they boost its all accuracy by 16.0\% and 31.4\% points with 5 and 20 exemplars, respectively.

\begin{table*}[t]
\begin{center}
\caption{CAMP achieves state-of-the-art results in most of the scenarios. We report known, novel, and all accuracy (\%) calculated over all classes after the last incremental learning step. We assume that task id is not known during the inference.
}
\resizebox{1.\textwidth}{!}{

\begin{tabular}{lccccccccccccccccc}
        \toprule
        \multirow{2}{*}
        {\textbf{Method}}
            & \multicolumn{1}{c}{Exemplars}
            & \multicolumn{3}{c}{CIFAR100}
            & \multicolumn{3}{c}{Stanford Cars}
            & \multicolumn{3}{c}{CUB200}
            & \multicolumn{3}{c}{FGVC Aircraft}
            & \multicolumn{3}{c}{DomainNet}
        \\ \cmidrule(lr){3-5} \cmidrule(lr){6-8} \cmidrule(lr){9-11} \cmidrule(lr){12-14} \cmidrule(lr){15-17}
            & \multicolumn{1}{c}{per class}
            & \multicolumn{1}{c}{All}
            & \multicolumn{1}{c}{Known}
            & \multicolumn{1}{c}{Novel}
            & \multicolumn{1}{c}{All}
            & \multicolumn{1}{c}{Known}
            & \multicolumn{1}{c}{Novel}
            & \multicolumn{1}{c}{All}
            & \multicolumn{1}{c}{Known}
            & \multicolumn{1}{c}{Novel}
            & \multicolumn{1}{c}{All}
            & \multicolumn{1}{c}{Known}
            & \multicolumn{1}{c}{Novel}
            & \multicolumn{1}{c}{All}
            & \multicolumn{1}{c}{Known}
            & \multicolumn{1}{c}{Novel}
        \\ \midrule

GCD~\cite{vaze2022generalized} & 0 & 26.7 & 26.6 & 27.2 & 20.6 & 23.8 & 5.8 & 20.6 & 22.5 & 13.3 & 16.8 & 20.0 & 4.4 & 24.7 & 23.7 & 28.5\\
GCD+EWC~\cite{kirkpatrick2017overcoming} & 0 & 26.9 & 28.7 & 19.7 & 30.4 & 35.3 & 8.7 & 50.3 & 53.3 & 38.1 & 25.3 & 28.2 & 13.8 & 33.1 & 37.5 & 16.8 \\
GCD+FD & 0 & 36.6 & 40.3 & 21.6 & 27.9 & 31.5 & 11.8 & 42.0 & 45.7 & 27.2 & 28.2 & 30.6 & 18.6 & 32.7 & 34.4 & \textbf{39.7}\\
MetaGCD~\cite{wu2023metagcd} & 0 & 35.8 & 40.0 & 19.1 & 23.6 & 25.6 & 14.4 & 42.0 & 44.6 & 31.4 & 28.6 & 31.0 & 19.0 & 30.8 & 34.2 & 37.0\\
SimGCD~\cite{wen2023parametric} & 0 & 31.2 & 31.9 & 28.5 & 21.3 & 30.1 & 23.9 & 35.7 & 36.6 & 31.8 & 22.0 & 22.3 & 20.8 & 36.5 & 36.4 & 35.0\\
CAMP & 0 & \textbf{52.1} & \textbf{55.7} & \textbf{37.8}
 & \textbf{48.8} & \textbf{52.8} & \textbf{31.0} & \textbf{58.9} & \textbf{62.6} & \textbf{44.2} & \textbf{39.9} & \textbf{42.0} & \textbf{31.5} & \textbf{36.7} & \textbf{39.2} & 29.7 \\

\midrule
GCD+ER\cite{riemer2018learning} & 5 & 31.7 & 32.9 & 27.0 & 32.4 & 38.1 & 6.5 & 36.6 & 42.4 & 13.7 & 36.6 & 42.4 & 13.7 & 34.2 & 38.4 & 34.2\\
PA~\cite{kim2023proxy} & 5 & 35.9 & 40.8 & 16.2 & 42.8 & 48.1 & 18.7 & 50.1 & 55.7 & 27.4 & \textbf{40.5} & 43.8 & \textbf{27.6} & 36.0 & 43.1 & 12.4 \\
MetaGCD~\cite{wu2023metagcd} & 5 & 38.9 & 43.6 & 20.0 & 34.5 & 38.9 & 14.5 & 48.2 & 52.0 & 32.9 & 37.8 & 41.3 & 24.0 & 34.8 & 38.5 & 22.1\\
IGCD~\cite{zhao2023incremental} & 5 & 40.6 & 43.6 & 28.8 & 37.8 & 41.3 & 22.4 & 50.8 & 54.8 & 34.8 & 34.5 & 36.4 & 26.9 & \textbf{45.9} & 35.8 & \textbf{43.4}\\
CAMP & 5 & \textbf{52.7} & \textbf{56.1} & \textbf{39.1} & \textbf{50.3} & \textbf{55.0} & \textbf{29.1} & \textbf{61.1} & \textbf{65.1} & \textbf{45.0} & 40.3 & \textbf{44.6} & 23.0 & 45.1 & \textbf{48.8} & 29.6 \\

\midrule 
GCD+ER\cite{riemer2018learning} & 20 & 42.5 & 46.3 & 27.3 & 46.3 & 54.0 & 11.5 & 52.0 & 59.9 & 20.1 & 48.9 & 56.5 & 17.5 & 46.4 & 53.7 & 20.4\\
PA~\cite{kim2023proxy} & 20 & 37.6 & 43.1 & 15.7 & 46.7 & 53.0 & 18.7 & 53.1 & 58.7 & 30.4 & 45.6 & 50.5 & 26.0 & 38.2 & 42.0 & 25.3\\
MetaGCD~\cite{wu2023metagcd} & 20 & 43.6 & 47.3 & 28.6 & 47.9 & 51.5 & 31.8 & 55.8 & 56.9 & 42.2 & 46.5 & 49.7 & 33.9 & 42.4 & 47.6 & 25.8\\
IGCD~\cite{zhao2023incremental} & 20 & 51.8 & 57.3 & 29.6 & 57.6 & 63.9 & 29.7 & 64.0 & 69.0 & 43.4 &  49.5 & 52.6 & \textbf{37.2} & 51.1 & 58.3 & 22.0\\
CAMP & 20 & \textbf{58.0} & \textbf{63.6} & \textbf{35.7} & \textbf{60.5} & \textbf{66.5} & \textbf{33.6} & \textbf{66.1} & \textbf{71.6} & \textbf{43.9} & \textbf{50.8} & \textbf{56.7} & 27.0 & \textbf{56.9} & \textbf{64.3} & \textbf{26.6} \\

\bottomrule
\end{tabular}

}

\label{tab:main_results}
\end{center}
\end{table*}

\begin{figure}[t]
    \centering
    \includegraphics[width=0.99\textwidth]{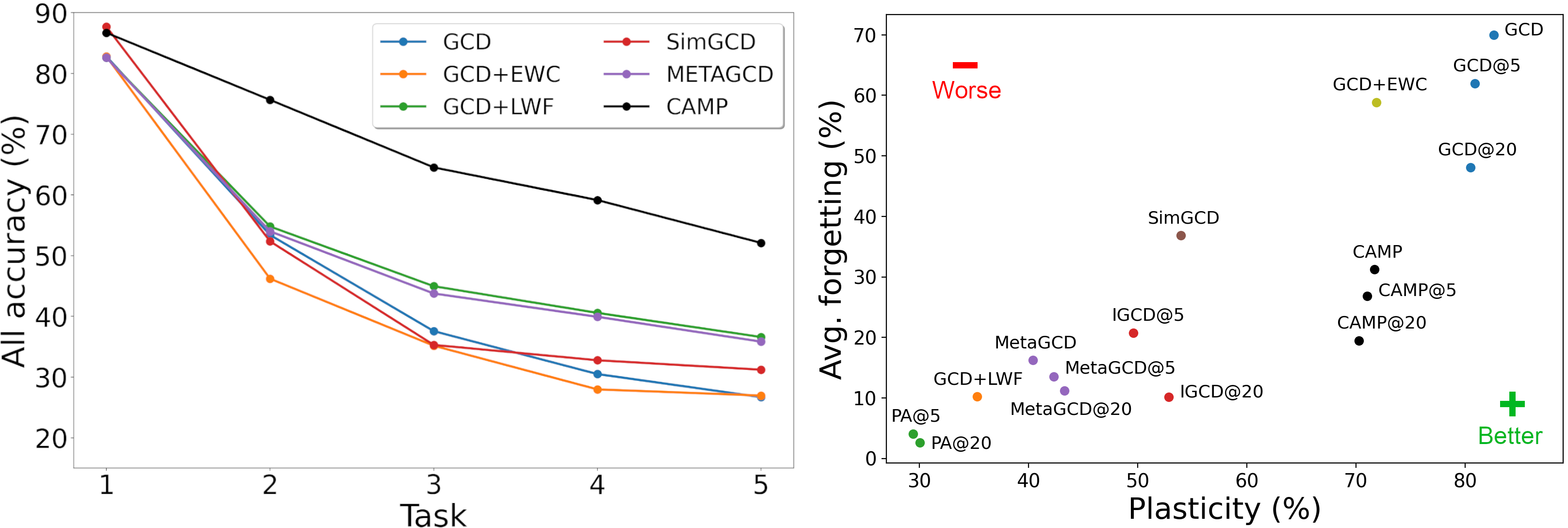}
    \caption{Results of GCCD methods on CIFAR100 during the continual learning session of five tasks. We report average accuracy after each task (\textbf{left}); plasticity and forgetting (\textbf{right}). CAMP achieves better accuracy and  than competitors.}
    \label{fig:acc_curves}
\end{figure}

To better examine how the accuracy changes in incremental steps, we plot averages of all accuracy after each task in Fig.~\ref{fig:acc_curves} (left). CAMP achieves the highest accuracy from the second task, and the results are consistent with and without exemplars. Next, we analyze the plasticity and forgetting of all methods in Fig.~\ref{fig:acc_curves} (right). GCD is the most plastic method. However, it has the highest forgetting as it has no form of regularization. On the other hand, IGCD is the most stable as it achieves the lowest forgetting. CAMP has good plasticity and forgetting, allowing it to achieve state-of-the-art results. We can also observe that exemplars prevent methods like GCD, IGCD, PA, and CAMP from forgetting. That is intuitive as the replay buffer allows them to recall the past categories' data. Interestingly, each method forms its own cluster in presented plots, informing that they have different but specific characteristics for continual learning scenarios.

\noindent\textbf{Class Incremental Learning results}
Class incremental learning~\cite{masana2022class} is a special case of GCCD, where each data sample is labeled, and all classes are known. We test CAMP against exemplar-free state-of-the-art methods and train the feature extractor (ResNet-18) from scratch. We train methods for $N$ equal tasks on CIFAR100, Domainnet, and Imagenet-Subset. For training CAMP we remove the self-supervised component from the loss function as all data is labeled. For baselines, we utilize popular exemplar-free CIL methods using their implementations in well-established benchmarks (FACIL~\cite{masana2022class}, PyCIL~\cite{zhou2023pycil}) and set their hyperparameters to default. We implement CAMP in FACIL and will publish the code upon the acceptance of the manuscript. For details, please refer to Supplementary Material B. We report average incremental accuracy~\cite{masana2022class}, which is the average of task agnostic accuracies measured after each task on tasks seen so far. We provide results in Tab.~\ref{tab:cil}.

CAMP outperforms baselines in terms of average incremental accuracy. On CIFAR100 it is better than the second best method - FeTrIL~\cite{petit2023fetril} by 6.5\%, 10.4\%, 4.3\% for 5, 10, 20 tasks respectively. FeTrIL freezes the model after the first task, what decreases its ability to adapt to new data. CAMP does not, thus allowing for better plasticity. Results are consistent on DomainNet and ImageNet-Subset. These results show that CAMP translates well to Class Incremental Learning scenario, therefore combining projected distillation and category adaptation is a promising technique in the field of Continual Learning.

\begin{table*}[t]
\begin{center}

\caption{Exemplar-free class incremental results for equal task split. We train all the methods from scratch and report task agnostic average incremental accuracy (\%). 
}
\label{tab:cil}
\resizebox{0.85\textwidth}{!}{
\begin{tabular}{@{\kern0.5em}lcccccccccccc@{\kern0.5em}}
        \toprule
        \multirow{2}{*}
        {\textbf{CIL Method}}
            & \multicolumn{3}{c}{CIFAR-100}
            && \multicolumn{3}{c}{DomainNet}
            && \multicolumn{3}{c}{ImageNet-Subset}
        \\ \cmidrule(lr){2-4} \cmidrule(lr){5-8} \cmidrule(lr){9-12} 
            & \multicolumn{1}{c}{\textit{N}=5}
            & \multicolumn{1}{c}{\textit{N}=10}
            & \multicolumn{1}{c}{\textit{N}=20}
            &
            & \multicolumn{1}{c}{\textit{N}=6}
            & \multicolumn{1}{c}{\textit{N}=12}
            & \multicolumn{1}{c}{\textit{N}=24}
            &
            & \multicolumn{1}{c}{\textit{N}=5}
            & \multicolumn{1}{c}{\textit{N}=10}
            & \multicolumn{1}{c}{\textit{N}=20}
        \\ \midrule

        \addlinespace[0.25em]
Fine-tune & 40.6 & 26.4 & 17.1 && 27.7 & 17.9 & 14.8 &&  41.6 & 27.4 & 18.6 \\
GCD~\cite{vaze2022generalized} & 40.1 & 25.2 & 16.5 &&  28.2 & 18.2 & 15.3 && 42.0 & 26.1 & 18.3 \\
EWC~\cite{kirkpatrick2017overcoming} & 52.9 & 37.8 & 21.0 &&  30.0 & 19.2 & 15.7 && 43.7 & 29.8 & 19.1 \\
LwF~\cite{li2017learning} & 49.4 & 47.0 & 38.5 &&  30.5 & 20.9 & 15.1 && 54.4 & 32.3  & 33.7\\   
PASS~\cite{zhu2021pass} & 62.2 & 52.0 & 41.4 &&  33.6 & 25.9 & 14.9 && 63.6 & 53.9 & 36.0\\ 
IL2A~\cite{zhu2021class} & - & 43.5 & 28.3 &&  31.3 & 20.7 & 18.2 && - & - & -\\ 
SSRE~\cite{zhu2022self} & 57.0 & 44.2 & 32.1 &&  34.9 & 33.2 & 24.0 && 56.9 & 45.0 & 32.9\\
FeTrIL~\cite{petit2023fetril} & 58.5 & 46.3 & 38.7 &&  36.8 & 33.5 & 27.5 && 62.9 & 58.7  & 43.2\\
CAMP  & \textbf{65.0} & \textbf{56.7} & \textbf{43.0} &&  \textbf{43.6} & \textbf{36.6} & \textbf{27.9} && \textbf{73.1} & \textbf{59.9}  & \textbf{50.2}\\ 
\hline 
Joint & \multicolumn{3}{c}{71.4} &&  65.1 & 63.7 & 69.3 && \multicolumn{3}{c}{81.5}\\
\hline
\end{tabular}
}
\end{center}
\end{table*}

\subsection{Method Analysis}
\label{sec:method_analysis}
\noindent\textbf{How do distillers and adapters affect performance?}
We verify the influence of different network architectures of adapters and distillers on the accuracy achieved by CAMP. For this purpose, we evaluate three types of adapter networks $\psi$: linear, two connected layers with batch-norm and ReLU activation after the first layer (MLP) as in~\cite{chen2020simple}, and three layers following~\cite{sariyildiz2023no} (T-ReX). We also test a semantic drift compensation~\cite{yu2020semantic} (SDC) method to compare adaptation with different, non-learnable methods. It utilizes a vector field to estimate the drift of prototypes in the latent space, and here, we use it to adapt centroids. As a distiller network $\phi$, we utilize popular feature distillation~\cite{Joseph2022NovelClassDisco ,roy2022class, kim2023proxy} technique and three networks of the same architecture as for adapters (Linear, MLP, T-ReX).

\begin{figure}
  \begin{minipage}[l]{.49\linewidth}
    \centering
    \includegraphics[width=0.99\linewidth]{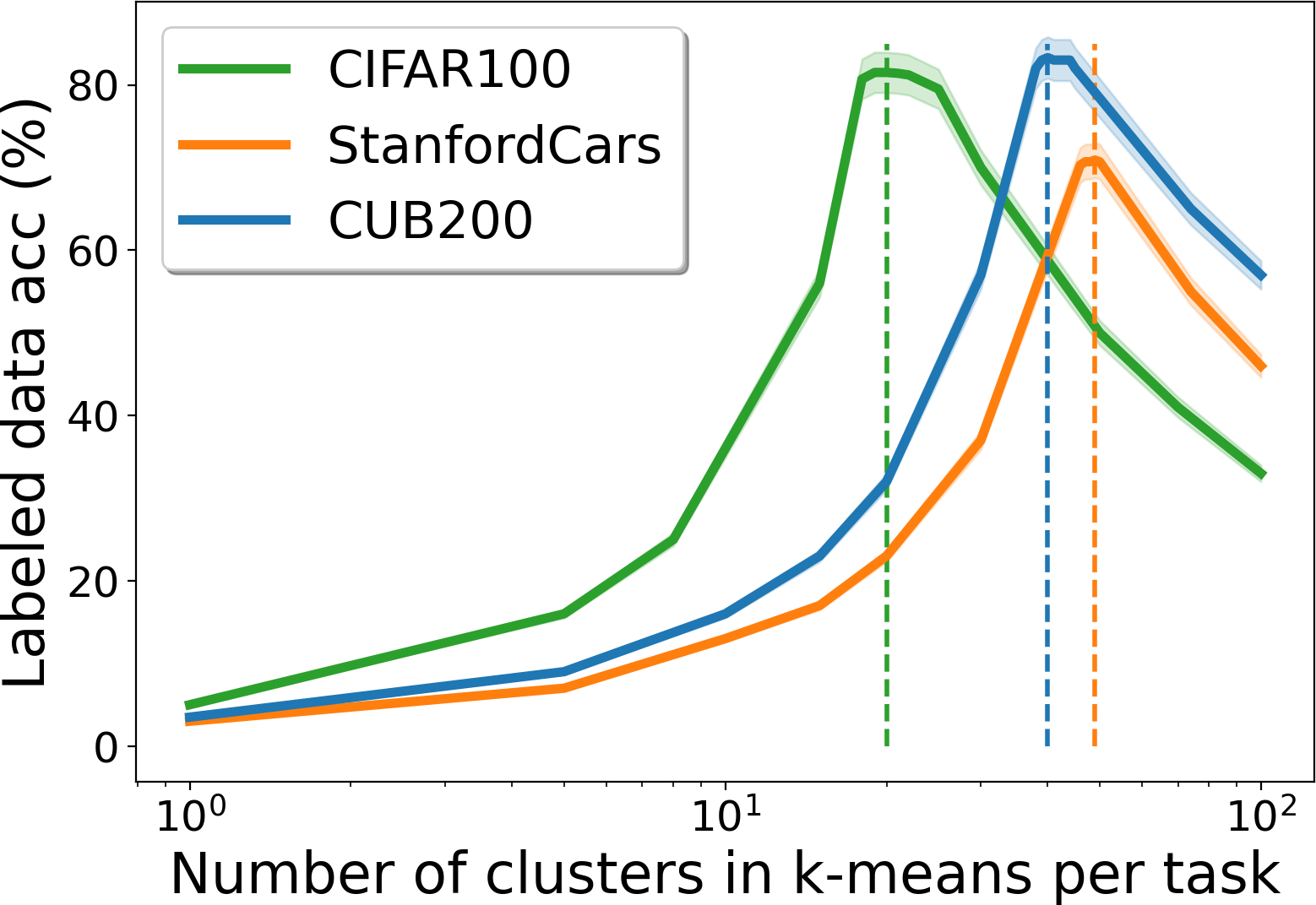}
    \caption{Results for estimation of number of categories in tasks. We utilize an elbow method following vanilla GCD method. Accuracy calculated on labeled data peaks around ground truth number of clusters.\label{fig:estimating}}
  \end{minipage}
  \hfill
  \begin{minipage}[r]{.49\linewidth}
    \centering
    \includegraphics[width=0.99\linewidth]{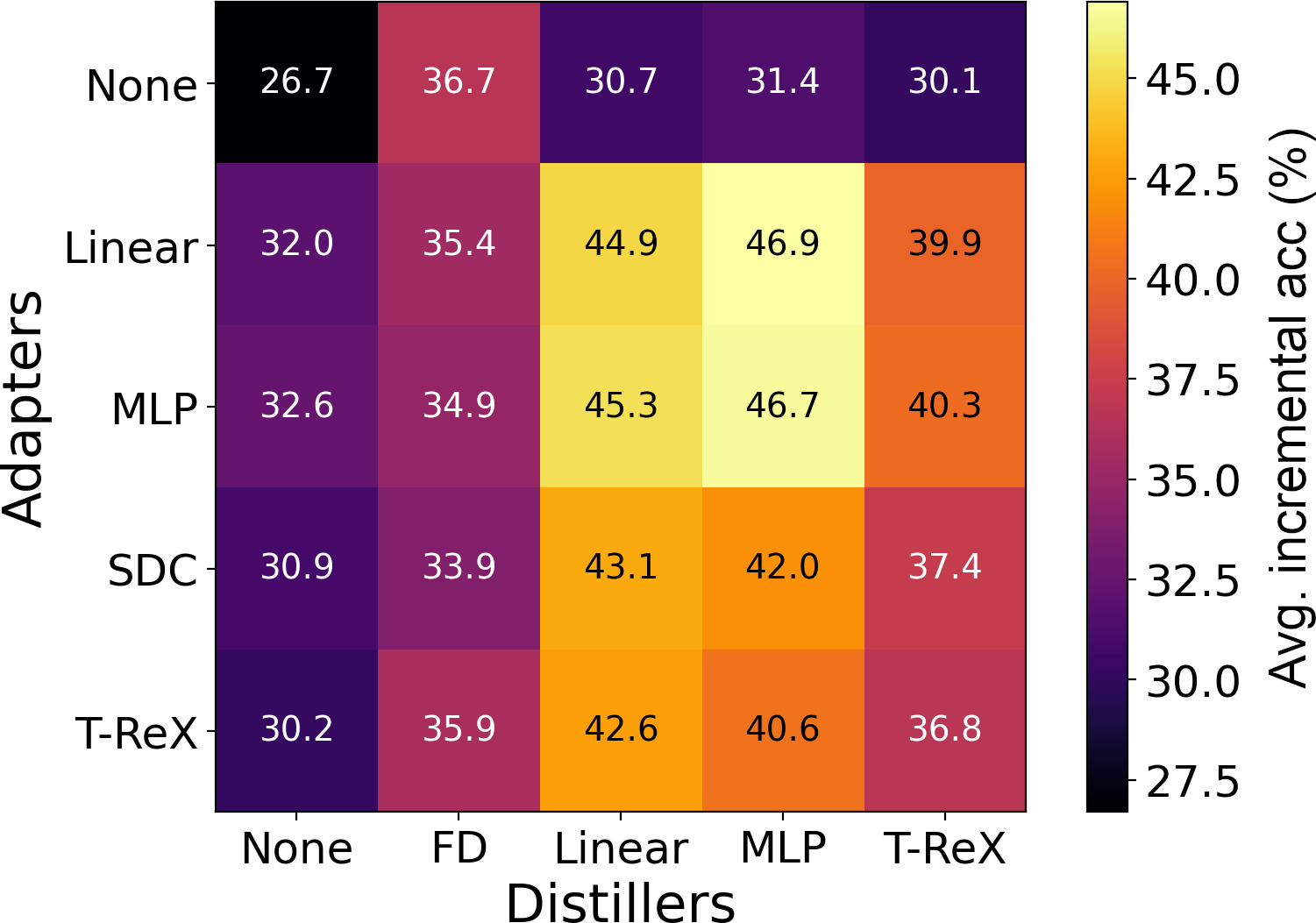}
    \caption{Performance of CAMP for different adapting ($\psi$) and distilling ($\phi$) networks. It achieves the best results when MLP distiller is combined with Linear or MLP adapter. \label{fig:adapters-distillers}}

  \end{minipage}

\end{figure}

We provide results for CIFAR100 in Fig.~\ref{fig:adapters-distillers}. It is best to combine MLP distiller and linear adapter, which justifies the design choice for CAMP. Utilizing only MLP distiller improves base results by 4.7\% points. However, combining MLP and linear adapter improves results by 20.2\%, which is a significant gain. We also achieve 12\% better accuracy than combination inspired by\cite{iscen2020memory} (MLP adapter and FD distiller). This shows that combining projected distillation with adaptation is crucial for achieving good performance. Interestingly, all adaptation methods improve no-adaptation results if FD distiller is not used. That suggests that FD changes old representation in a harder-to-predict way than other distillers. 

\begin{figure}[ht!]
    \centering
    \includegraphics[width=0.99\textwidth]{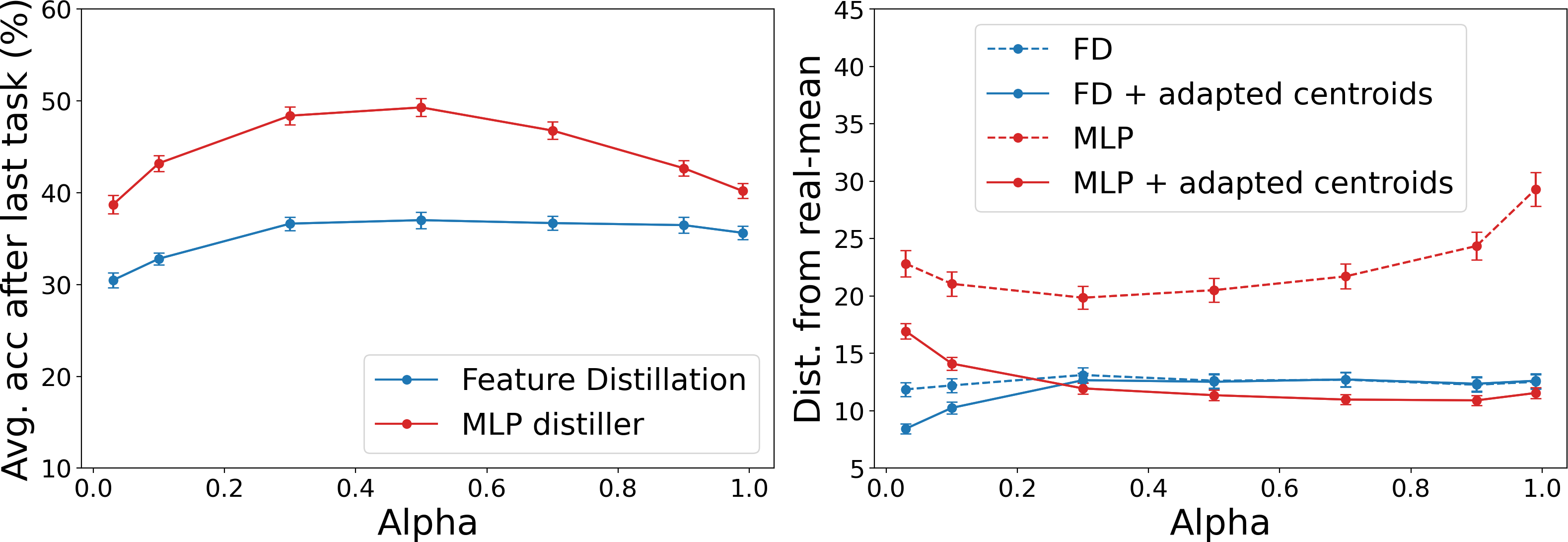}
    \caption{MLP projector relaxes drifts of class distributions (measured with distance from memorized to ground-truth class centroids) compared to feature distillation. Additionally, it enables better adaptation of centroids, leading to increased accuracy.} 
    \label{fig:mlp-vs-fd}
\end{figure}

\noindent\textbf{Adaptation with feature distillation or projected distillation?}
We examine the impact of Feature Distillation (FD) and distillation with an MLP projector on centroid adaptation for $\alpha$ parameter value in range [0.03, 0.99] (plasticity-stability trade-off). We present the results in Fig.~\ref{fig:mlp-vs-fd}. We can see that without adaptation, FD imposes more substantial constraints on the drift of old centroids than MLP, as the distance between real and memorized centroids is lower (right). However, after adapting centroids this distance is much lower for MLP, proving that MLP facilitates drift prediction. Moreover, the adaptation method does not work well for FD with $\alpha \geq 0.4$, as the rigid features regularization minimizes the gain from centroid adaptation. The accuracy when using MLP is much better than FD, which proves the design choice of the distilling method.

\noindent\textbf{Ablation study}
We perform ablation study on CUB200 and FGVCAircraft datasets. We present results in Tab.~\ref{tab:ablation}. We test CAMP without: $\mathcal{L}_{SL}$, $\mathcal{L}_{SSL}$, knowledge distillation through the projector loss ($\mathcal{L}_{KD}$) and without centroid adaptation. We can see that having all of these elements is crucial for obtaining good accuracy on known and novel classes (62.6\% and 44.2\% respectively on CUB200). Interestingly, without the $\mathcal{L}_{SSL}$  CAMP achieves 2.0\% points better accuracy for known classes but 2.8\%  points lower accuracy for novel classes.

\begin{table}[h]
\begin{center}
\caption{Ablation study of CAMP.}
 \label{tab:ablation}
\resizebox{0.8\textwidth}{!}{

\begin{tabular}{@{\kern0.5em}cccccccc@{\kern0.5em}}
        \toprule
 &  &  & 
 & \multicolumn{2}{c}{CUB200}  & \multicolumn{2}{c}{FGVCAircraft} 
 \\ \cline{5-6} \cline{7-8} 
\multirow{-2}{*}{ $\mathcal{L}_{SL}$} 
& \multirow{-2}{*}{$\mathcal{L}_{SSL}$} 
& \multirow{-2}{*}{$\mathcal{L}_{KD}$}
& \multirow{-2}{*}{Cent. adapt.}

& Known & Novel & Known & Novel
\\ \midrule
\XSolidBrush  & $\checkmark$ & $\checkmark$ & $\checkmark$  & 29.9$\pm$1.6 &12.1$\pm$1.1 & 19.0$\pm$0.9 & 18.8$\pm$1.9 \\
$\checkmark$ & \XSolidBrush  & $\checkmark$ & $\checkmark$  & 64.6$\pm$2.3 & 41.4$\pm$3.5 & 43.5$\pm$1.8 & 16.6$\pm$2.0 \\
$\checkmark$ & $\checkmark$ & \XSolidBrush & $\checkmark$  & 30.9$\pm$2.6 & 32.9$\pm$2.6 & 24.7$\pm$1.2 & 19.8$\pm$2.2 \\
$\checkmark$ & $\checkmark$ & $\checkmark$ & \XSolidBrush   & 36.5$\pm$1.9 & 28.7$\pm$1.7 & 29.6$\pm$1.7& 24.2$\pm$2.8\\
\hline
$\checkmark$ & $\checkmark$ & $\checkmark$ & $\checkmark$ & 62.6$\pm$2.4 & 44.2$\pm$3.7 & 42.0$\pm$1.9 & 31.5$\pm$3.5 \\
\hline
\end{tabular}
}
\end{center}
\end{table}

\noindent\textbf{Estimating number of categories in a task.} We estimate the number of novel categories in each task using the elbow method. We present results in Fig.~\ref{fig:estimating}. It is clear that accuracy on the labeled dataset peaks when the number of clusters for k-means algorithm is equal to the ground truth number of categories.

\begin{figure}
    \centering
    \includegraphics[width=1\textwidth]{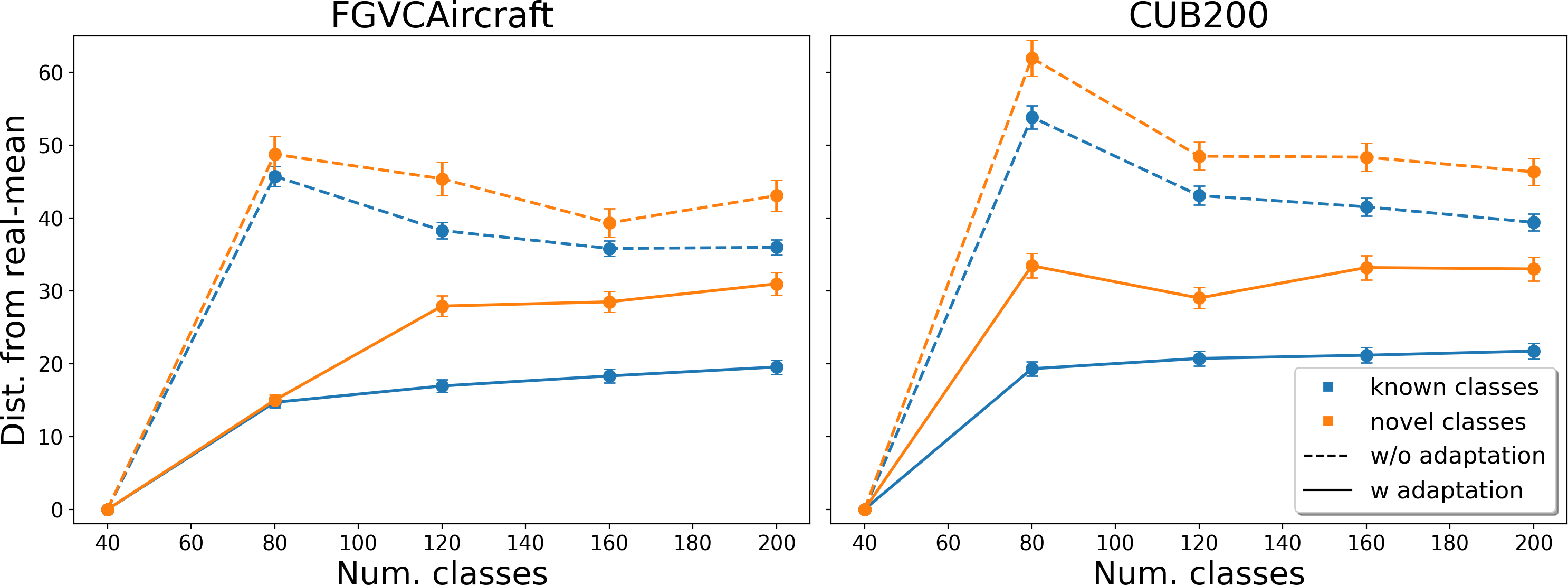}
    \caption{Distance from centroids estimated by CAMP to ground-truth centroids throughout sequential training. Novel categories tend to be further away from estimated positions than known ones. CAMP centroid adaptation improves these estimations in both cases.} 
    \label{fig:adapt}
\end{figure}

\noindent\textbf{Does category adaptation improve position of centroids?}
After each task, we measure the distance between ground truth and estimated centroids with and without our category adaptation method. We present results in Fig.~\ref{fig:adapt}. Interestingly, we can see that novel classes tend to be further away from their estimated positions than known ones. Our centroid adaptation method consistently improves estimations of real centroids, \eg after the last incremental step it decreases mismatch between positions of known classes by 44.8\% and novel classes by 28.8\% on CUB200.

\noindent

\begin{figure}
  \begin{minipage}{.49\linewidth}
  \centering
  \includegraphics[width=0.99\linewidth]{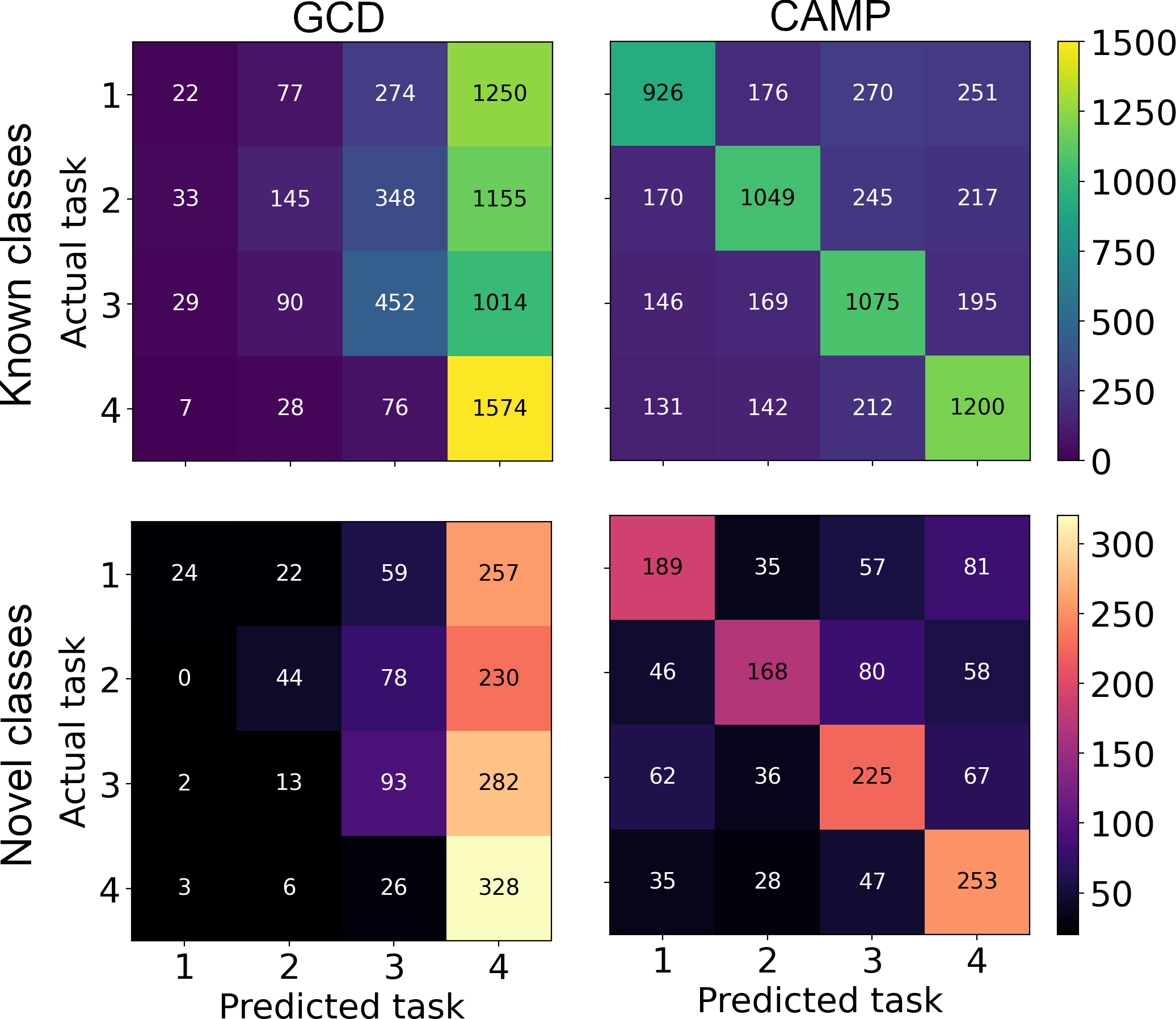}
      \caption{Vanilla GCD method is prone to task-recency bias as its predictions are skewed towards the last one. However, CAMP overcomes this problem and keeps predictions distributed across all tasks. }\label{fig:task-bias}

  \end{minipage}\hfill
  \begin{minipage}{.49\linewidth}
  \centering
  \includegraphics[width=0.99\linewidth]{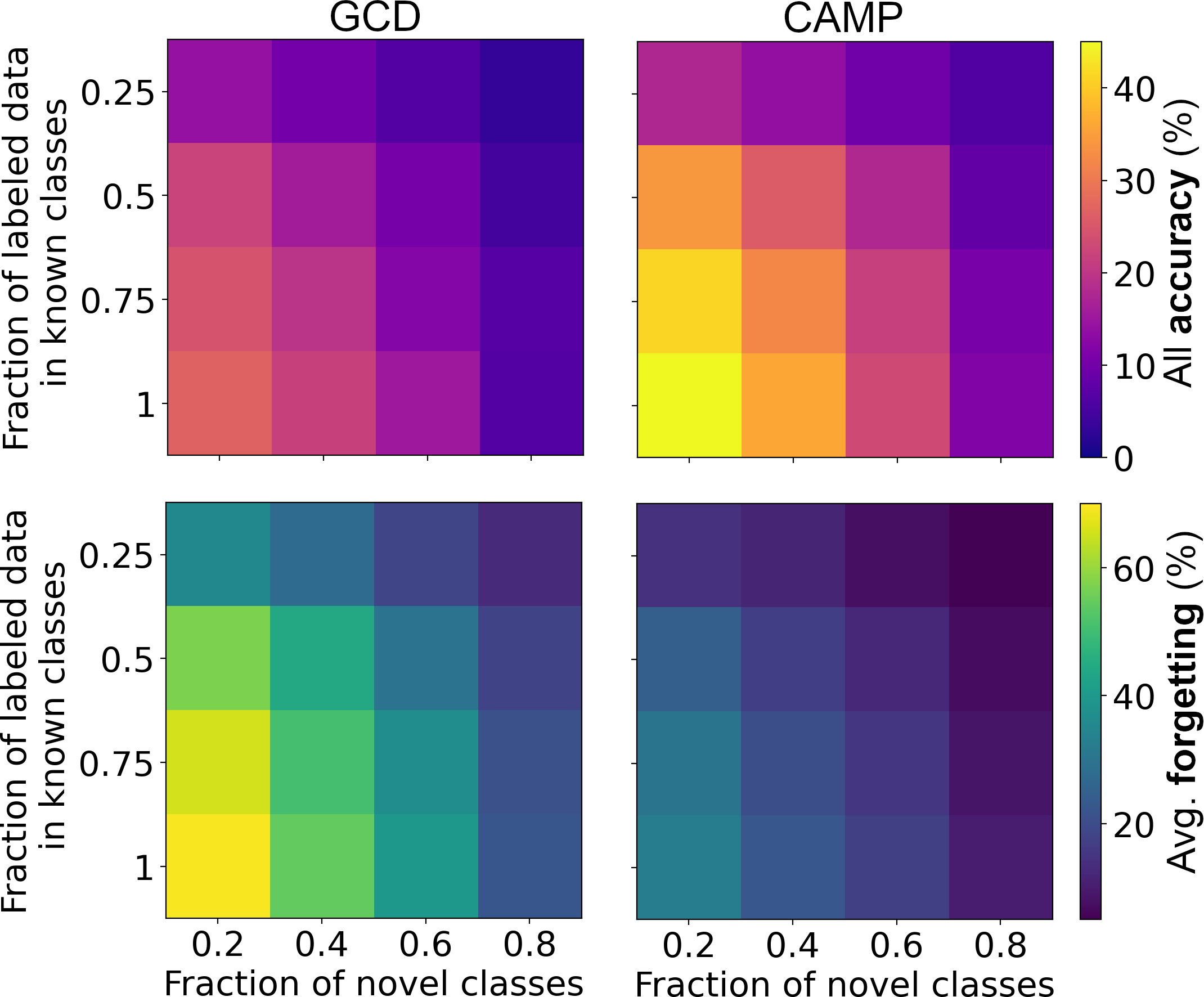}
      \caption{CAMP improves GCD in different continual learning scenarios in terms of average accuracy and forgetting. We evaluate methods for different numbers of novel classes in each task and different fractions of labeled known data on StanfordCars.}\label{fig:heatmaps}
  \end{minipage}\hfill
\end{figure}

\noindent\textbf{CAMP overcomes task-recency bias.}
To better understand the performance gap between GCD and CAMP, we check how these methods predict tasks after the last incremental step. We present a confusion matrix calculated for the StanfordCars dataset split into four tasks in Fig.~\ref{fig:task-bias}. As a fine-tuning method, GCD faces a task-recency bias~\cite{wu2019large} for known and novel classes, as it mainly classifies test samples into categories of the last task. On the contrary, CAMP does not present such skewed results prediction only towards the last task. Our method of projected distillation, combined with category adaptation, alleviates task-recency bias, which improves CAMP's performance.

\noindent
\textbf{CAMP vs. GCD under different setting conditions.}
We examine the impact of setting parameters (number of novel classes and fraction of labeled data) on results achieved by CAMP and GCD. We set novel to known ratios as 2:8, 4:6, 6:4, 8:2 and fraction of labeled data as 0.25, 0.5, 0.75, 1.0. We provide final all accuracy and average forgetting on StanfordCars in Fig.~\ref{fig:heatmaps}. We can observe that accuracy decreases when increasing the number of novel classes or when decreasing the amount of labeled data. Forgetting behaves similarly, which is understandable - with lower accuracy, there is less to forget. However, in each case, CAMP outperforms GCD, proving that our combination of distillation and category adaptation is robust to different scenarios.

\section{Summary}
In this work, we study an interplay between knowledge distillation techniques and centroid adaptation in the problem of GCCD and CIL. We find out that combining knowledge distillation through a neural projector followed by centroid adaptation is crucial for achieving high plasticity and low forgetting. Based on this finding, we propose a new method, dubbed CAMP. It utilizes projected knowledge distillation to regularize the network and prevent forgetting. Moreover, CAMP utilizes a centroid adaptation method to estimate better actual positions of centroids representing past categories. CAMP uses the nearest centroid classifier to classify data samples achieving state-of-the-art results in different scenarios.


\noindent
\textbf{Limitations.} The limitation of our work is that the centroid adaptation mechanism cannot be easily implemented for methods that do not perform centroid-based clustering to represent categories, \eg~\cite{zhao2023incremental, wen2023parametric}. Moreover, our method achieves better results for known classes, which can be limiting in use cases where novel classes play the leading role.

\section*{Acknowledgments}
Daniel Marczak is supported by National Centre of Science (NCN, Poland) Grant No. 2021/43/O/ST6/02482.
This research was partially funded by National Science Centre, Poland, grant no: 2020/39/B/ST6/01511, 2022/45/B/ST6/02817, and 2023/51/D/ST6/02846.
Bartłomiej Twardowski acknowledges the grant RYC2021-032765-I.
This paper has been supported by the Horizon Europe Programme (HORIZON-CL4-2022-HUMAN-02) under the project "ELIAS: European Lighthouse of AI for Sustainability", GA no. 101120237.
We gratefully acknowledge Polish high-performance computing infrastructure PLGrid (HPC Center: ACK Cyfronet AGH) for providing computer facilities and support within computational grant no. PLG/2023/016613.

\clearpage  

%
%
\bibliographystyle{splncs04}
\bibliography{main}

\title{Supplementary Material: Category Adaptation Meets Projected Distillation in Generalized Continual Category Discovery}

\titlerunning{Supplementary Material}

\author{
Grzegorz~Rypeść\thanks{Corresponding author, email: grzegorz.rypesc.dokt@pw.edu.pl}\inst{1,2}\orcidlink{0000-0001-8170-3282} \and
Daniel~Marczak\inst{1,2}\orcidlink{0000-0002-6352-9134} \and
Sebastian~Cygert\inst{1,3}\orcidlink{0000-0002-4763-8381} \and \\
Tomasz~Trzciński\inst{1,2,4}\orcidlink{0000-0002-1486-8906} \and
Bartłomiej~Twardowski\inst{1,5,6}\orcidlink{0000-0003-2117-8679}
}

\authorrunning{G.~Rypeść et al.}

\institute{
IDEAS~NCBR \and
Warsaw~University~of~Technology \and 
Gdańsk~University~of~Technology \and 
Tooploox \and
Autonomous~University of~Barcelona \and 
Computer~Vision~Center
}

\maketitle

\section{CAMP training loss in detail}
\label{appendix:camp_details}

In this section, we describe the details of CAMP method, more specifically -- representation learning during phase 1. Representation learning in CAMP consists of contrastive and entropy-based learning, which we describe below and conclude in a total loss function for CAMP at the end.

\textbf{Contrastive learning}
We use unsupervised and supervised contrastive losses, namely SimCLR~\cite{chen2020simple} and SupCon~\cite{khosla2021supervised}.

We calculate SimCLR~\cite{chen2020simple} loss as:
\begin{equation}
\label{eq:simclr}
\mathcal{L}_{SimCLR} = - \frac{1}{|B|} \sum_{i \in B}\log \frac{\exp \left(\boldsymbol{h}_i^\top \boldsymbol{h}_i^{\prime} / \tau\right)}{\sum_i^{i \neq n} \exp \left(\boldsymbol{h}_i^\top \boldsymbol{h}_n^{\prime} / \tau\right)} \,,
\end{equation}
where $x_i$ and $x_{i}'$ are two views (random augmentations) of the same image in a mini-batch $B$, $\mathbf{h}_i = g(\mathcal{F}(x_i))$, and $\tau$ is a temperature value. $\mathcal{F}$ is the feature extractor, and $g$ is a multi-layer perceptron (MLP) projection head used in SimCLR method.

For data where labels are available ($T^t_L$) we can simply use supervised learning. Specifically, we employ SupCon~\cite{khosla2021supervised} loss:
\begin{small}
\begin{equation}
\label{eq:supcon}
\medmuskip=2mu
\thickmuskip=3mu
\renewcommand\arraystretch{1.5}
\mathcal{L}_{SupCon} = - \frac{1}{|B_L|} \sum_{i \in B_L}\frac{1}{|\mathcal{N}_i|} \sum_{q \in \mathcal{N}_i}\log \frac{\exp \left(\boldsymbol{h}_i^\top \boldsymbol{h}_q^{\prime} / \tau\right)}{\sum_i^{i \neq n} \exp \left(\boldsymbol{h}_i^\top \boldsymbol{h}_n^{\prime} / \tau\right)} \,,
\end{equation}
\end{small}
where $B_{L}$ corresponds to the labeled subset of $B$ and $\mathcal{N}_i$ is the set of indices of other images that have the same label as $x_i$. We use the same projector as in SimCLR to produce representations $\mathbf{h}_i = g(\mathcal{F}(x_i))$ and the same temperature parameter $\tau$.

\textbf{Entropy based learning}

We also utilize two additional loss functions for training the feature extractor following~\cite{wen2023parametric} and applied in the self-distillation~\cite{caron2021emerging,assran2022masked} fashion. We noticed that this improved results achieved by CAMP even though we do not utilize parametrical classifier like~\cite{wen2023parametric}.
Formally, for a total number of categories $K$, we randomly initialize a set of prototypes $\mathcal{C}=\{\boldsymbol{c}_1, \dots, \boldsymbol{c}_K\}$, each representing one category.
During training, we calculate the soft label for each augmented view $\boldsymbol{x}_i$ by performing softmax on cosine similarity between the hidden feature $\boldsymbol{z}_i=\mathcal{F}(x_i)$ and the prototypes $\mathcal{C}$ scaled by $1/\tau_s$:
\begin{equation}
\boldsymbol{p}_{i}^{(k)}=\frac{\exp \left(\frac{1}{\tau_s}(\boldsymbol{z}_i/||\boldsymbol{z}_i||_2)^\top (\boldsymbol{c}_k / ||\boldsymbol{c}_k||_2)\right)}{\sum_{k^\prime} \exp \left(\frac{1}{\tau_s}(\boldsymbol{z}_i / ||\boldsymbol{z}_i||_2)^\top (\boldsymbol{c}_{k^\prime} / ||\boldsymbol{c}_{k^\prime}||_2)\right)} \,,
\end{equation}
and the soft pseudo-label $\boldsymbol{q}_i^\prime$ is produced by another view $\boldsymbol{x}_i$ with a sharper temperature $\tau_t$ in a similar way.
The loss functions are then cross-entropy loss $\ell(\boldsymbol{q}^{\prime}, \boldsymbol{p}) = -\sum_{k} {\boldsymbol{q}^{\prime(k)}}\log{\boldsymbol{p}}^{(k)}$ between the predictions and pseudo-labels:
\begin{equation}
    \mathcal{L}_{pseudo} = \frac{1}{|B|}\sum_{i \in B}\ell(\boldsymbol{q}_i^\prime, \boldsymbol{p}_i) - \varepsilon H(\overline{\boldsymbol{p}}) ,
\end{equation}
or known labels:
\begin{equation}
     \mathcal{L}_{CE} = \frac{1}{|B_L|}\sum_{i \in B_L}\ell(\boldsymbol{y}_i, \boldsymbol{p}_i)
     ,
\end{equation}
where $\boldsymbol{y}_i$ denotes the one-hot label of $\boldsymbol{x}_i$.
For the unsupervised objective, we additionally adopt a mean-entropy maximisation regulariser~\cite{assran2022masked}.
Here, $\overline{\boldsymbol{p}} = \frac{1}{2|B|}\sum_{i \in B}\left( \boldsymbol{p}_i+\boldsymbol{p}_i^\prime \right)$ denotes the mean prediction of a batch, and the entropy $H(\overline{\boldsymbol{p}}) = -\sum_k\overline{\boldsymbol{p}}^{(k)}\log\overline{\boldsymbol{p}}^{(k)}$.

\textbf{Total loss for CAMP}
On all data provided in a given task: $\mathcal{X}^t = \mathcal{X}^t_U + \mathcal{X}^t_L$, we calculate the following loss as $\mathcal{L}_{SSL} = \mathcal{L}_{SimCLR} + \mathcal{L}_{pseudo}$

Loss that we calculate only on the labeled data $\mathcal{X}^t_L$ is equal to $\mathcal{L}_{SL} = \mathcal{L}_{SupCon} + \mathcal{L}_{CE}$
Finally, the total loss function for CAMP is then equal to:
\begin{equation}
        \mathcal{L}_{CAMP}=(1-\alpha)((1-\beta)\mathcal{L}_{SSL} + \beta \mathcal{L}_{SL}) + \alpha \mathcal{L}_{KD},
\end{equation} where $\alpha, \beta \in [0,1]$ are hyperparameters defining contribution of regularization and supervision respectively and  $\mathcal{L}_{KD}$ is knowledge distillation loss calculated using a distiller and defined in Section 3 (main paper). In experimental section we set $\alpha$ to 0.5 or 0.1 (when exemplars are present) and $\beta$ to 0.35. In case of Class Incremental Learning scenario we set $\alpha$ to 0.9.

\section{Experimental setup - details}
\subsection{Generalized Continual Class Discovery}
In order to fairly compare existing methods, which often were created for a very specific continual scenario, we evaluate them in a Generalized Continual Category Discovery framework (GCCD). GCCD consists of an arbitrary number of disjoint tasks and can include exemplars. Each task consists of labeled and unlabeled data from known and novel classes. We extend well-established category incremental setting~\cite{rebuffi2017icarl, buzzega2020dark} by adding unlabeled data and novel classes to each task. We formally define it in Section 3 of the main paper. We present differences between GCCD and other setting in Tab.\ref{tab:setting_characteristic}.

\begin{table*}
  \centering
  \scalebox{0.65}{
\begin{tabular}{lccccc}\hline
\multirow{4}{*}{\diagbox{Setting}{Characteristics}} 
& & Partially & Novel & Sequence & Arbitrary\\
& Methods & labeled & classes & of disjoint & number\\
& & classes & & tasks & of tasks\\
\midrule
Class incremental learning & LwF~\cite{li2017learning}, EWC~\cite{kirkpatrick2017overcoming}, iCaRL\cite{rebuffi2017icarl}, DER~\cite{buzzega2020dark}& $\times$ & $\times$ & \checkmark & \checkmark\\
Self-supervised continual learning & CaSSLe~\cite{fini2022cassle}, PFR~\cite{gomezvilla2022continually}, LUMP~\cite{madaan2022lump}, POCON~\cite{gomezvilla2024plasticity} & $\times$ & $\checkmark$ & \checkmark & \checkmark\\
Semi-supervised continual learning & NNCSL~\cite{kang2022soft}, CCIC~\cite{boschini2021continual}, ORDisCo~\cite{wang2021ordisco} & \checkmark & $\times$ & \checkmark & \checkmark\\
Generalized category discovery & GCD~\cite{vaze2022generalized}, SimGCD~\cite{wen2023parametric} & \checkmark & \checkmark & $\times$ & $\times$ \\
Incremental generalized category discovery & IGCD~\cite{zhao2023incremental} & \checkmark & \checkmark & $\times$ & \checkmark\\
Continual generalized  category discovery & PA~\cite{kim2023proxy} & \checkmark & \checkmark & \checkmark & $\times$\\
\textbf{GCCD (ours)} & & \checkmark & \checkmark & \checkmark & \checkmark\\
\midrule
\end{tabular}}
  \caption{Generalized Continual Category Discovery is the most general setting. It includes partial labels, contrary to supervised and self-supervised learning, and requires discovering novel (unlabeled) classes, contrary to supervised and semi-supervised learning. Moreover, it works on a sequence of disjoint tasks, contrary to Incremental Generalized Category Discovery, which assumes that unlabeled data samples become labeled in the next task. Finally, unlike Continual Generalized Category Discovery, GCCD is not limited to only a single category incremental step.}
  \label{tab:setting_characteristic}
\end{table*}

Overall, the proposed setting in this paper is a generalization of the previous ones as we learn both known and novel classes in each of the learning stages and allow for partially-labeled classes, thus we make no distinction between the initial and subsequent learning stages. This scenario holds in many real-life applications, mostly when data comes sequentially and there are insufficient resources to label all images. Additionally, our experiments focus on equal split tasks (so without a sizeable first task) on many incremental steps and investigate the effect of different setting parameters, such as the fraction of novel classes or labeled data proportion, on models performance. For most GCCD experiments (Tab. 1, Fig.6, Fig.9 of the main body) we assume that methods know the number of novel classes due to simplicity purposes. We will publish the code of the framework and our method upon the acceptance of the manuscript. We present data splits which we used for evaluation of GCCD protocol in~\ref{tab:datasets}.

\begin{table}
\centering
\scalebox{0.9}{
        \begin{tabular}{lccccc}
        \toprule
        Dataset & N & Known & Novel & Lab. known (\%)\\
        \midrule
        CIFAR100 & 5  & 16 & 4 & 50\%\\
        Stanford Cars & 4 & 40 & 9 & 50\%\\
        CUB200 & 5 & 32 & 8 & 50\%\\
        Aircrafts & 5 & 16 & 4 & 50\%\\
        DomainNet & 6 & 8 & 2 & 50\%\\
        \bottomrule
        \end{tabular}}
        \caption{Datasets we utilized in experiments, their splits and characteristics.}
        
    \label{tab:datasets}
\end{table}

\noindent\textbf{Exemplars}
In our experimental setting we utilize exemplars only for known classes as only they are given labels. We randomly choose which exemplars to store in the buffer. For each past, known class we store the same amount of exemplars.

\subsection{Class Incremental Learning}
Here, we provide setup details for results presented in Tab.2 of the main body of the paper. Class Incremental Learning\cite{li2017learning, masana2022class} is a special case of GCCD, where each data sample is labeled and there are no novel classes. To combat forgetting, typical approaches utilize exemplars\cite{rebuffi2017icarl, iscen2020memory}, expendable architectures~\cite{yan2021dynamically, rypesc2024divide} or regularization~\cite{li2017learning, yu2020semantic, kirkpatrick2017overcoming} techniques. We compare CAMP to regularization baselines that consider a single neural network with constant number of parameters and no exemplars.

For the simplest baselines we utilize fine-tuning (no technique to combat forgetting) and joint that trains the network on the whole dataset. Then, we utilize LwF\cite{li2017learning} which utilizes knowledge distillation, EWC~\cite{kirkpatrick2017overcoming} that regularizes weights of the neural network and PASS~\cite{zhu2021pass} as a prototype augmentation technique. Additionally, we use IL2A~\cite{zhu2021class} and~\cite{petit2023fetril}. For all this methods we run their implementations in FACIL\cite{masana2022class} and if the implementation is not available, we use PyCIL\cite{zhou2023pycil}. For all methods we use default hyperparameters and the same augmentations.

To compare CAMP to baselines, we utilize three commonly used benchmark datasets in the field of Continual Learning (CL): CIFAR-100~\cite{krizhevsky2009learning} (100 classes), ImageNet-Subset~\cite{deng2009imagenet} (100 classes) and DomainNet~\cite{peng2019moment} (345 classes, from 6 domains). DomainNet contains categories of different domains, allowing us to measure models' adaptability to new data distributions. We create each task with a subset of classes from a single domain, so the domain changes between tasks. We split datasets to $N$ equal tasks.

We train CAMP and CIL baseline methods from scratch. We compare all approaches with standard CIL evaluations using the classification accuracies after each task, and \emph{average incremental accuracy}, which is the average of those accuracies~\cite{rebuffi2017icarl}.

\section{Baseline methods details}
In the following, we describe details of training feature extractors $\mathcal{F}$ of baseline methods.

\textbf{GCD}~\cite{vaze2022generalized} combines contrastive unsupervised loss from Eq.~\ref{eq:simclr} and supervised loss from Eq.~\ref{eq:supcon} to train $\mathcal{F}^t$ on task $t$.  Combined, the total $\mathcal{L}_{GCD}$ loss equals:
\begin{equation}
    \mathcal{L}_{GCD} = (1-\beta)\mathcal{L}_{SimCLR} + \beta \mathcal{L}_{SupCon},
\end{equation}
where $\beta \in (0,1)$ is a weighing parameter and is equal to 0.35 in our experiments following the original work~\cite{vaze2022generalized}.

\textbf{GCD+FD} 
GCD method was not designed to tackle continual scenarios as it suffers from catastrophic forgetting. In order to adapt it for the GCCD we follow most distillation-based continual learning methods~\cite{li2017learning, gomezvilla2022continually, fini2022cassle}. We freeze previously trained feature extractor $\mathcal{F}^{t-1}$ and regularize currently trained feature extractor $\mathcal{F}^t$ with the outputs of $\mathcal{F}^{t-1}$:
\begin{equation}
\mathcal{L}_{FD} = \frac{1}{|B|}\sum_{i \in B} || \mathcal{F}^{t}(x_i) - \mathcal{F}^{t-1}(x_i) ||^2,
\end{equation}
This form of distillation does not require labels and all the data from the current task can be used for a regularization.

The final loss function for feature extractor training is defined as follows:
\begin{equation}
    \mathcal{L}_{GCD+FD} = (1-\alpha)\mathcal{L}_{GCD} + \alpha \mathcal{L}_{FD},
\end{equation}
where $\alpha \in [0,1]$ is a hyperparameter defining the contribution of regularization.

\textbf{GCD+EWC}
In this baseline method we improve GCD by enforcing Elastic Weight Consolidation  regularization on $\mathcal{F}$ parameters using $\lambda$ parameter as described in \cite{kirkpatrick2017overcoming}. We additionally add a linear head for training and change $\mathcal{L}_{SupCon}$ to cross entropy loss. In our experiments we set $\lambda$ to 5000 following the original work.

\textbf{SimGCD}~\cite{wen2023parametric} improves training $\mathcal{F}^t$ over GCD by adding two additional loss functions: popular cross entropy loss for labeled data which improves clustering capabilities and adapted mean-entropy maximisation regularisation~\cite{assran2022masked} applied to all data present in the task. The loss is equal to 
\begin{equation}
        \mathcal{L}_{SimGCD}=((1-\beta)\mathcal{L}_{SSL} + \beta \mathcal{L}_{SL})
\end{equation}

\textbf{IGCD}~\cite{zhao2023incremental} uses the same loss function as SimGCD to train the feature extractor. However, it adapts SimGCD for continual settings by adding a replay buffer that helps to mitigate forgetting and provides for support sample selection. In each incremental step a random subset of data samples of each class in the task is added to the buffer. For classification IGCD utilizes Soft-Nearest-Neighbor classifier.

\textbf{PA}~\cite{kim2023proxy} utilizes proxy anchors to train $\mathcal{F}$ and a replay buffer to fight the forgetting. Proxy anchor loss is defined as:
\begin{equation}
\resizebox{0.8\columnwidth}{!}{
    $
    \begin{aligned}
        \mathcal{L}_{SupPA}&=\frac{1}{\lvert{P^{0^+}}\rvert}\sum_{p\in{P^{0^+}}}\log\bigg(1+\sum_{z\in{Z^{0^+}_p}}{e^{-\mu(s(z, p)-\delta)}}\bigg)\\        
        &+\frac{1}{\lvert{P^{0}}\rvert}\sum_{p\in{P^0}}\log\bigg(1+\sum_{z\in{Z^{0^-}_p}}{e^{\mu(s(z, p)+\delta)}}\bigg)
    \end{aligned}
    $
}
\label{eq:proxy_loss}
\end{equation}
\noindent where $\mu>0$ is a scaling factor and $\delta>0$ is a margin which we set to 32 and 0.1 respectively, following the original work.
Here, the function $s(\cdot{,}\cdot)$ denotes the cosine similarity score. $P^{0^+}$ represents same class PAs(\eg{negative}) in the batch. Each proxy $p$ divides the set of embedding vector $Z^0$ as $Z^{0^+}_p$ and $Z^{0^-}_p=Z^0-Z^{0^+}_p$. $Z^{0^+}_p$ denotes the same class embedding points with the proxy anchor $p$. The goal of the first term in the equation is to pull $p$ and its dissimilar but hard positive data together, while the last term is to push $p$ and its similar but hard negatives apart. The proxies are incrementally added in each task.

Following the original work~\cite{kim2023proxy}, the total loss for PA method for training $\mathcal{F}$ is equal to:
\begin{equation}
    \mathcal{L}_{PA} = \mathcal{L}_{SupPA} + \mathcal{L}_{KD}
\end{equation}

\section{Additional experiments}

\noindent\textbf{Comparison to baselines}

\begin{figure*}[ht!]
    \centering
    \includegraphics[width=1\textwidth]{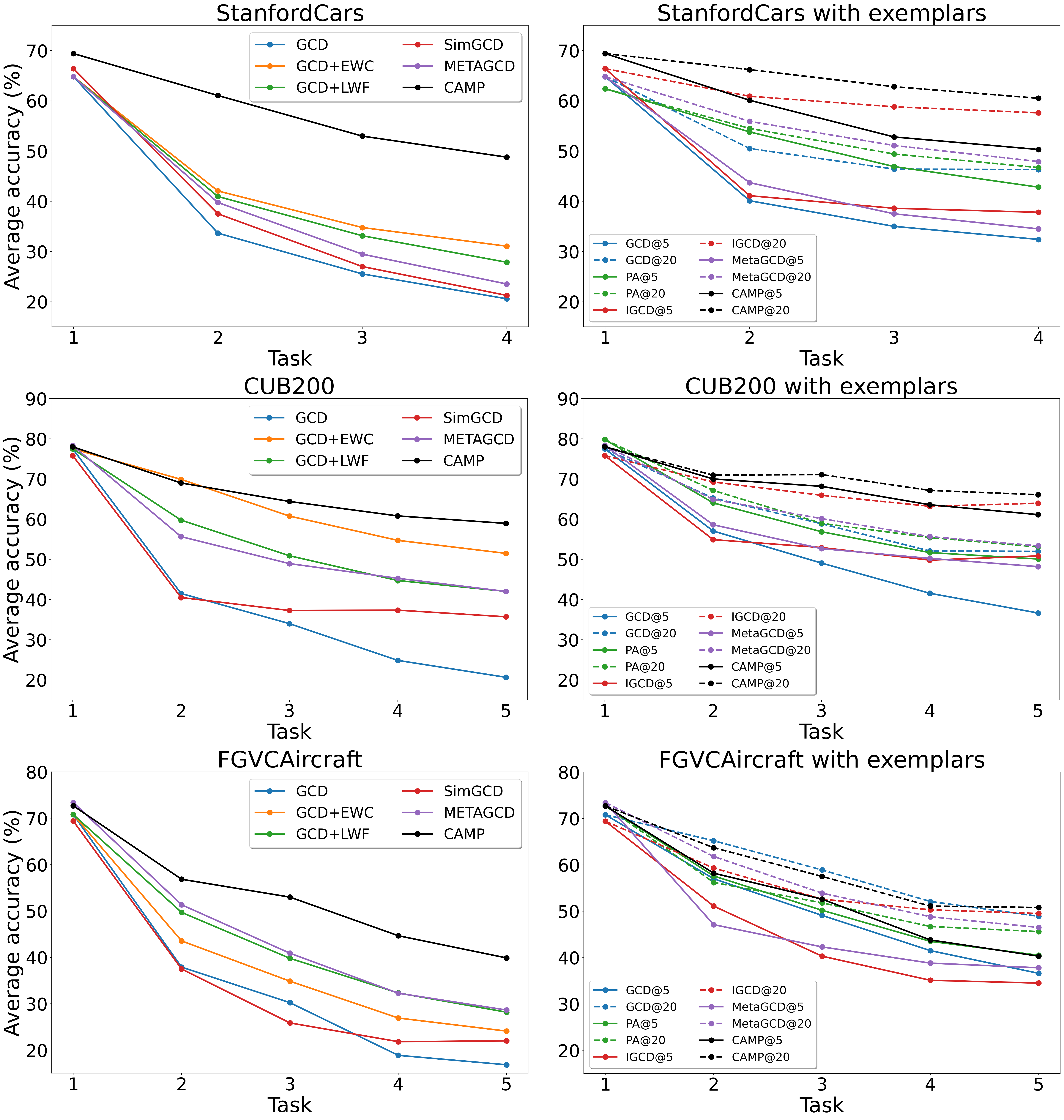}
    \caption{Average accuracy after each task on three datasets. CAMP achieves the best accuracy after most of tasks.} 
    \label{appdx:fig:accuracies}
\end{figure*}

We provide additional average accuracy after each task plots in Fig.~\ref{appdx:fig:accuracies}. They represent results achieved by exemplar and exemplar-free methods on StanfordCars, CUB200, and FGVC Aircraft. CAMP achieves the best final accuracy, and results are consistent on all datasets. CAMP also achieves the best average accuracy after most of the tasks. However, on FGVCAircraft CAMP with 20 exemplars, is worse than GCD, with 20 exemplars after the second, third, and fourth tasks.

\begin{figure}[ht]
 \hfill
  \begin{minipage}[l]{.49\linewidth}
    \centering
    \includegraphics[width=0.99\linewidth]{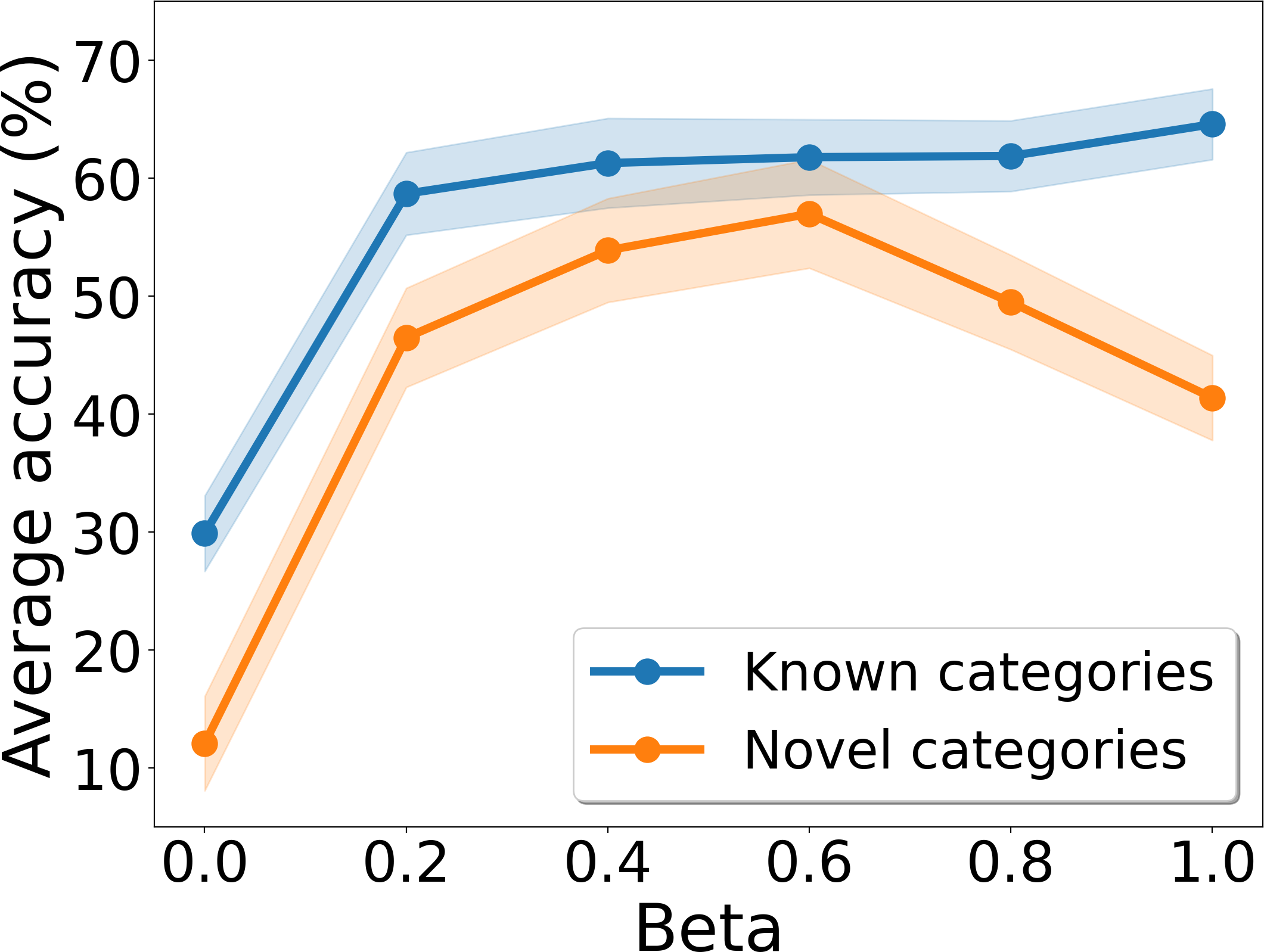}
        \caption{Impact of $\beta$ hyperparametrer on known and novel accuracy achieved on CUB200. \label{fig:beta}}
  \end{minipage}
  \hfill
  \begin{minipage}[r]{.49\linewidth}
    \centering
    \includegraphics[width=0.99\linewidth]{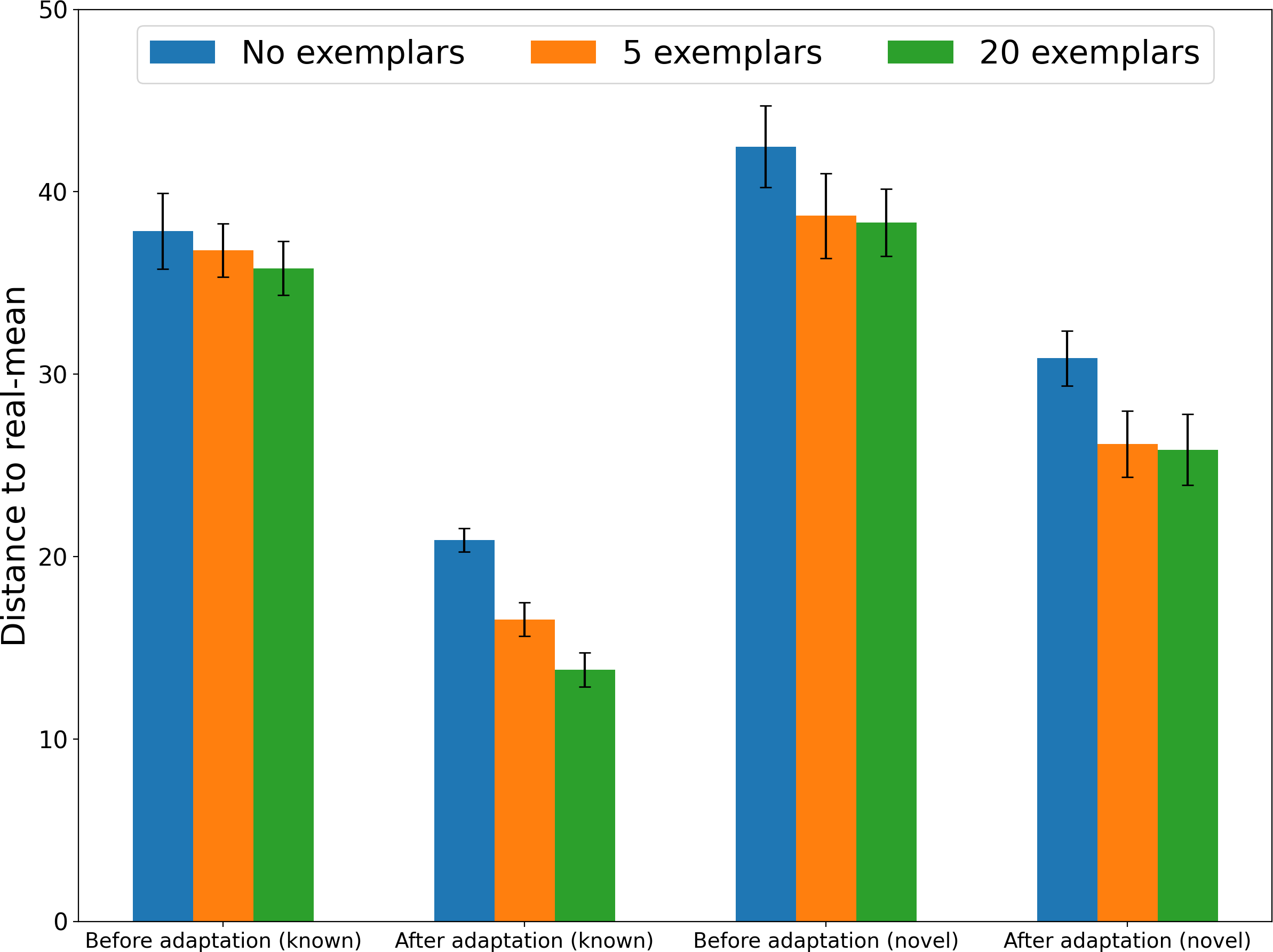}
    \caption{Distance to real-mean before and after adaptation for 0, 5, 20 exemplars on CUB200. \label{fig:adaptation}}
  \end{minipage}
\end{figure}

\noindent\textbf{Impact of $\beta$ hyperparameter}
We verify the impact of $beta$ hyperparameter (trade-off between SL and SSL losses) on CUB200. We measure average accuracy for all classes for $\beta$ equal to $0, 0.2, 0.4, 0.6, 0.8$ and $1.0$ and present results in \ref{fig:beta}. CAMP achieves very low results for $\beta=0$ as SL loss is not utilized in this case. Interestingly, accuracy for novel categories drops for $\beta > 0.6$ showcasing that the SSL part is crucial in obtaining good results for novel categories.

\noindent\textbf{Impact of number of exemplars on centroid adaptation}
We verify how the number of exemplars (0, 5, 20 per each category) influences distance of memorized centroid to the real category centroid (denoted as distance to real-mean). We plot the results for CUB200 dataset split into 5 tasks in Fig.~\ref{fig:adaptation}. We measure the distance to the real-mean before and after performing the centroid adaptation. Intuitively, the more exemplars are available, the better is the centroid estimation. This results are consistent for known and novel categories.

\noindent\textbf{Distillers vs adapters for known and novel classes}

We verify the impact of using different distillation functions and different adapters for CAMP on StanfordCars dataset split into four tasks. We utilize the same setup as in 4.4 (main body). We provide final all, known, novel accuracies in Fig.~\ref{appdx:tab:distiller_vs_adapter_cars}. The combination of MLP distiller and Linear adapter achieves 38.1\%, 43.2\%, and 15.0\% on all, known and novel categories, respectively, which is the best result. This shows, that such combination improves results for known and novel classes. The results are consistent with Fig.5 (main paper). That proves the design choice of our architecture.

\begin{table}[t]
  \centering
  \scalebox{0.9}{
\begin{tabular}{lccccc}\hline
\multirow{3}{*}{\diagbox{Adapter}{Distiller}} 
&&&&\\
& None & FD & Linear & MLP & T-ReX\\
&&&&\\
\midrule
None & 20.6 & 28.3 & 25.2 & 22.5 & 21.3\\
Linear & 25.6 & 28.5 & \underline{35.5} & \textbf{38.1} & 26.5 \\
MLP & 26.6 & 26.0 & 34.4 & 33.7 & 23.1\\
T-ReX~\cite{sariyildiz2023no} & 24.3 & 23.9 & 33.4 & 28.9 & 21.8\\
SDC~\cite{yu2020semantic} & 23.8 & 28.1 & 29.4 & 31.0 & 23.8\\
\bottomrule
\\
\end{tabular}}

  \centering
  \scalebox{0.9}{
\begin{tabular}{lccccc}\hline
\multirow{3}{*}{\diagbox{Adapter}{Distiller}} 
&&&&\\
& None & FD & Linear & MLP & T-ReX\\
&&&&\\
\midrule
None & 23.9 & 32.1 & 28.9 & 25.8 & 24.5 \\
Linear & 29.4 & 32.5 & \underline{40.2} & \textbf{43.2} & 30.1\\
MLP & 30.3 & 30.1 & 39.5 & 38.8 & 26.5 \\
T-ReX~\cite{sariyildiz2023no} & 28.5 & 27.8 & 38.5 & 32.9 & 24.9 \\
SDC~\cite{yu2020semantic} & 28.0 & 32.4 & 33.3 & 35.5 & 27.0\\
\bottomrule
\\
\end{tabular}}

  \centering
  \scalebox{0.9}{
\begin{tabular}{lccccc}\hline
\multirow{3}{*}{\diagbox{Adapter}{Distiller}} 
&&&&\\
& None & FD & Linear & MLP & T-ReX\\
&&&&\\
\midrule
None & 5.8 & 11.3 & 8.2 & 7.7 & 7.0 \\
Linear & 8.6 & 10.2 & \underline{14.3} & \textbf{15.0} & 10.5  \\
MLP & 9.7 & 7.4 & 11.4 & 11.1 & 8.2 \\
T-ReX~\cite{sariyildiz2023no} & 7.2 & 6.3 & 10.5 & 10.7 & 8.0 \\
SDC~\cite{yu2020semantic} & 6.9 & 10.3 & 9.8 & 11.0 & 9.6\\
\bottomrule

\end{tabular}}
  \caption{Impact of different adapters and distiller on our method on StanfordCars. We report all (top), known (middle) and novel (bottom) accuracy after the last task. Combination of MLP distiller and Linear adapter achieves the best results on all types of categories proving the design choice of CAMP.}
    \label{appdx:tab:distiller_vs_adapter_cars}
\end{table}

\noindent\textbf{Extended analysis of latent spaces in CAMP}

\begin{figure*}[ht!]
    \centering
    \includegraphics[width=0.98\textwidth]{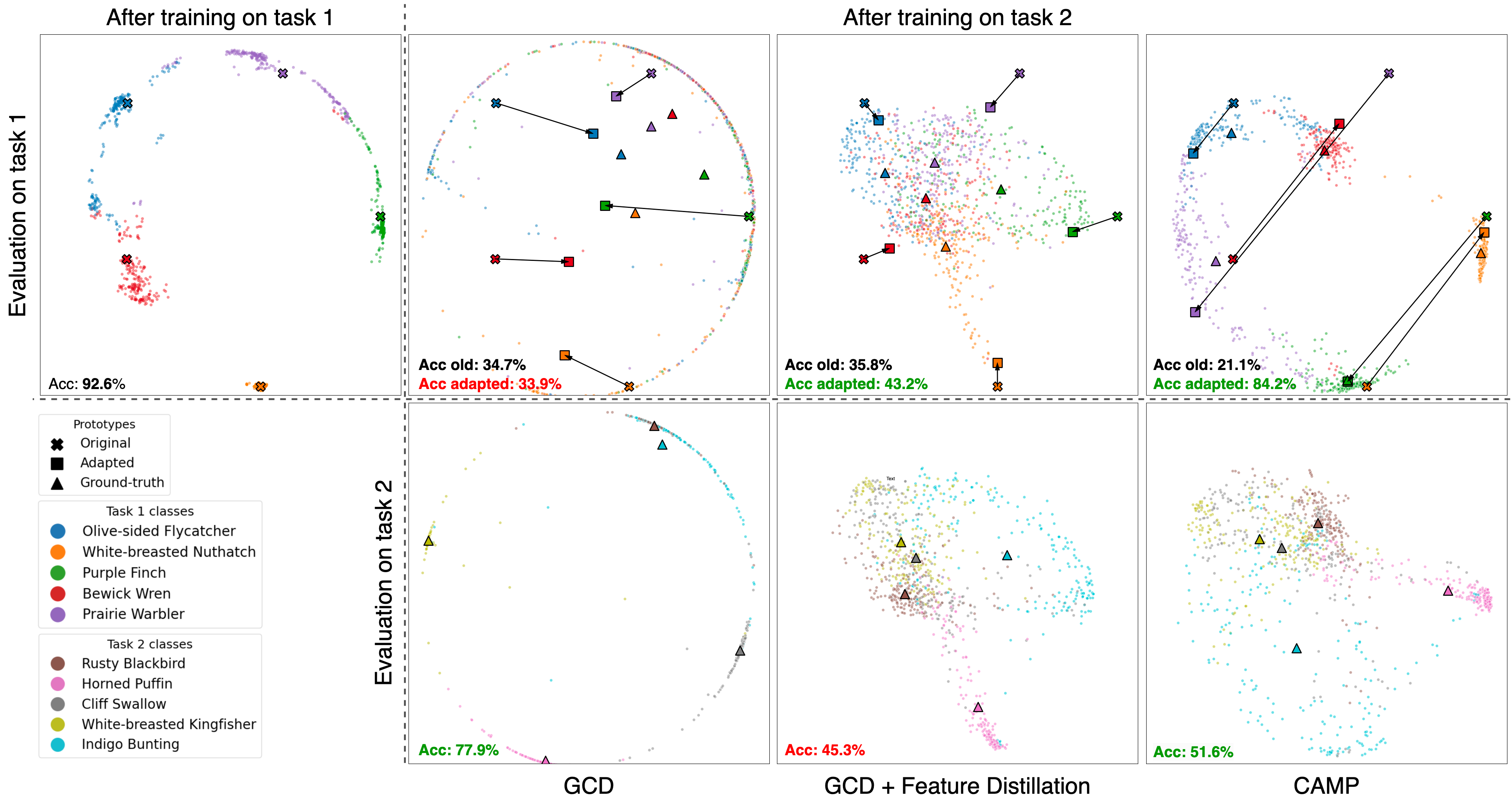}
    \caption{CAMP utilizes a projected knowledge distillation that results in: (1) predictable drift in latent space that is revertible via centroids adaptation and leads to high performance on the first task and (2) high plasticity of the model and its ability to learn new tasks leading to high performance on the second task. Vanilla GCD fails to prevent forgetting. However, GCD with feature distillation reduces forgetting but diminishes the ability to learn new tasks. We report the nearest centroid classification accuracy. On the first task, we report accuracy using stored prototypes (Acc old) and adapted prototypes (Acc adapted).} 
    \label{appdx:fig:drift_viz}
\end{figure*}

In Fig.~\ref{appdx:fig:drift_viz}, we present an analysis of latent spaces in different GCCD methods. We observe that using no distillation (GCD) leads to the best performance on the second task but the worst performance on the first task, as the model is optimized to fit the new data without regard for the past task. A rigid distillation (GCD + Feature Distillation) helps prevent forgetting but hinders learning new tasks. Moreover, we can see that feature distillation leads to the overlap between tasks. Our CAMP method uses projected knowledge distillation that enhances the ability to learn new tasks and learn representations that overlap less with those of old categories.

\clearpage  

%
%

\end{document}